\documentclass[11pt,oneside, a4paper]{scrartcl}
\usepackage[english]{babel}
\usepackage[utf8]{inputenc}
\usepackage[T1]{fontenc}
\usepackage{amssymb}
\usepackage{amsmath}
\usepackage{amsthm}
\usepackage{amsfonts}
\usepackage[pdftex]{graphicx}
\usepackage{enumitem}
\usepackage{url}
\usepackage{epstopdf}
\usepackage{todonotes}
\usepackage{mathtools}
\usepackage[sort&compress,numbers]{natbib}

\graphicspath{{Figures/}}
\usepackage[abs]{overpic}
\newcommand{\SE}[1][3]{\mathrm{SE}(#1)}

\newcommand{\PS}{\mathbb{P}^7}

\newcommand{\etalchar}[1]{$^{#1}$}

\usepackage{url}

\usepackage{hhline}

\title{Analysis of a 3-\underline{R}UU Parallel Manipulator}

  \author{Thomas Stigger\thanks{ \{thomas.stigger, johannes.siegele, daniel.scharler, martin.pfurner, manfred.husty \} @uibk.ac.at}, Johannes Siegele, Daniel F. Scharler, \\ Martin Pfurner, and Manfred L. Husty}
\date{Unit of Geometry and CAD, University of Innsbruck, \\6020 Innsbruck, Austria}

\begin{document}
 \maketitle
%\begin{frontmatter}
%

\begin{abstract}
  The aim of this paper is to give a detailed examination of the input and
  output singularities of a 3-\underline{R}UU parallel manipulator in the
  translational operation mode. This task is achieved by using algebraic
  constraint equations. For this type of manipulator a complete workspace
  representation in Study coordinates is presented after elimination of the
  input parameters. Both, input and output singularities are mapped into a Study
  subspace as well as into the joint space. Therewith a detailed singularity
  investigation of the translational operation mode of a 3-\underline{R}UU
  parallel manipulator is provided. This paper is an extended version of a
  previous publication. The addendum comprises the discovery of a possible
  transition between two operation modes as well as a self motion and an
  examination of another component of the output singularity surface, most of
  them for arbitrary design parameters.
  
\end{abstract}

% \begin{keyword}
% 3-\underline{R}UU \sep singularity analysis \sep singularities \sep workspace \sep algebraic
%  geometry \sep operation mode \sep self motion \sep Study space \sep joint space
% \end{keyword}

\section{Introduction}

There are many different approaches to perform the kinematic analysis of
parallel manipulators (PM). More recently dual quaternions have been used to
describe the group of Euclidean displacements $\SE$. Together with the use of
Study's kinematic mapping they have proven to be very successful to obtain
information about the global behavior of parallel manipulators. The aims of this
kinematic analysis are to derive algebraic constraint equations, to solve the
direct and inverse kinematics, to describe the complete workspace, operation
modes and all singularities (e.g.~\cite{HuPfuSchBru}).

There are numerous papers investigating workspace and singularities of parallel
manipulators especially of the Stewart-Gough platform e.g. by Dasgupta and T.S.
Mruthyunjaya~\cite{DASGUPTA200015}, Borr{\`a}s \emph{et
  al.}~\cite{10.1007/978-90-481-9262-5_45} or Nawratil~\cite{5524579,
  NAWRATIL20101851}. Also closely related manipulators were already investigated
for example the 4-\underline{R}UU in Amine \emph{et al.}~\cite{Amine2012}.

There are also numerous papers on lower degree of freedom parallel
manipulators, most of them use vector loop equations or screw theory
for the kinematic analysis. Analyzing the global kinematics of lower
degree of freedom parallel manipulators using algebraic constraint
equations one can explain the overall kinematic behavior,
e.g. determining different operation modes, transition between the
operation modes and all singularities of the manipulator (for
example~\cite{Schadlbauer2014}).

This paper is an extension of~\cite{Stigger2018b} which used results of
~\cite{Stigger2018} to provide an even more detailed singularity analysis of the
translational operation mode of a 3-\underline{R}UU parallel manipulator.
Furthermore, an algebraic representation of the workspace is derived, which
might prove to be very useful in several kinematic tasks like path planning. To
the best of the authors knowledge this is the first non-trivial lower degree of
freedom PM known with an input parameter free description of the workspace.
Section \ref{sec:1} recalls the manipulator's architecture and the algebraic
constraint equations that describe its motion capabilities. Section \ref{sec:2}
shows the algorithm to derive the workspace equations, which are only displayed
for the translational mode in this paper due to their length. In Section
\ref{sec:3} a thorough singularity analysis of the translational mode is given
comprising a detailed description of input and output singularities in the
kinematic image space (Study space) as well as in the joint space. The main part
of the extension comprises an investigation on constraint singularities and
possible transition between operation modes. Additionally, a profound
examination of the output singularities results in the discovery of a self
motion as well as a new component of the singularity surface. In contrast
to~\cite{Stigger2018b}, most of the results are derived for arbitrary design
parameters. The last section concludes the results.

\section{Manipulator Architecture and Constraint Equations}\label{sec:1}

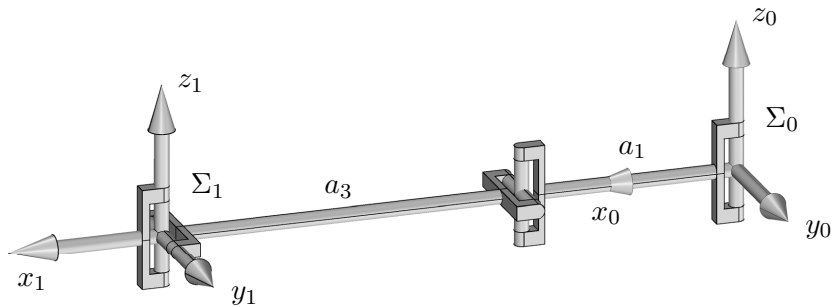
\begin{figure}[b!]
    \centering
    \begin{overpic}[scale=.38]%
        {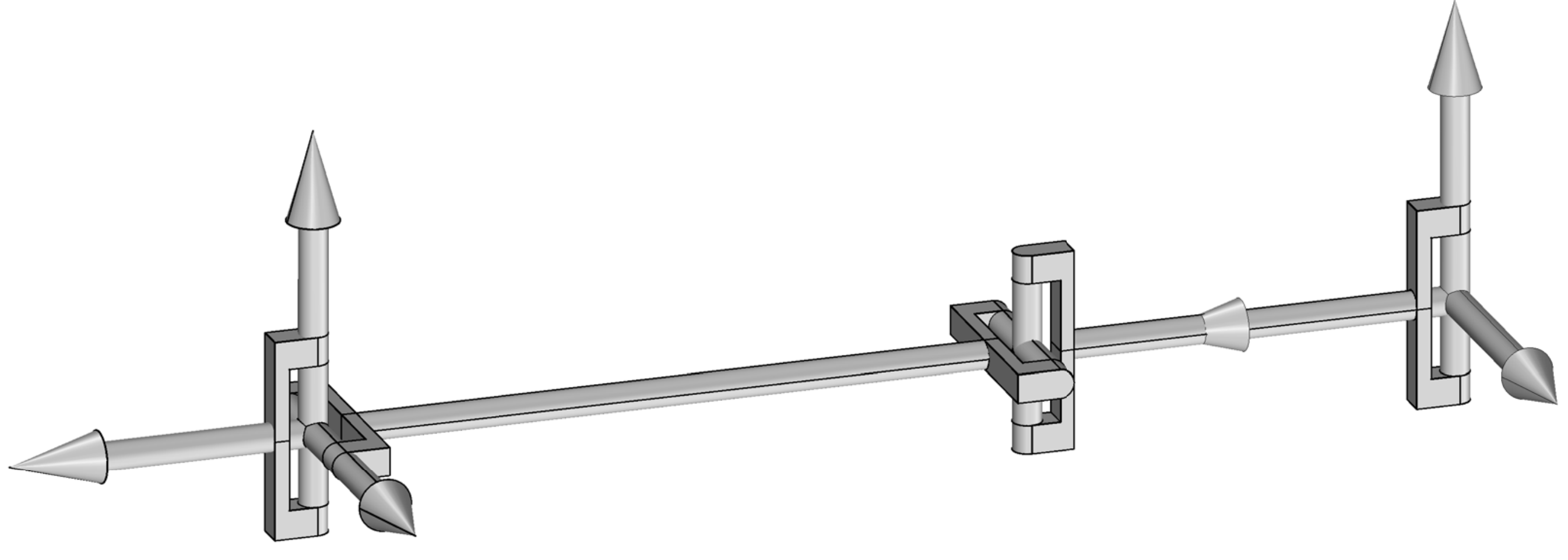}
        \put(285,65){$\Sigma_0$}
        \put(70,40){$\Sigma_1$}
        \put(230,55){$a_1$}
        \put(120,40){$a_3$}
        \put(220,30){$x_0$}
        \put(300,25){$y_0$}
        \put(280,105){$z_0$}
        \put(5,5){$x_1$}
        \put(85,0){$y_1$}
        \put(65,80){$z_1$}
    \end{overpic}\\
    \caption{A \underline{R}UU limb}\label{fig:ruulimb}
\end{figure}

\begin{figure}[ht!]
    \centering
    \begin{overpic}[scale=.38]%
        {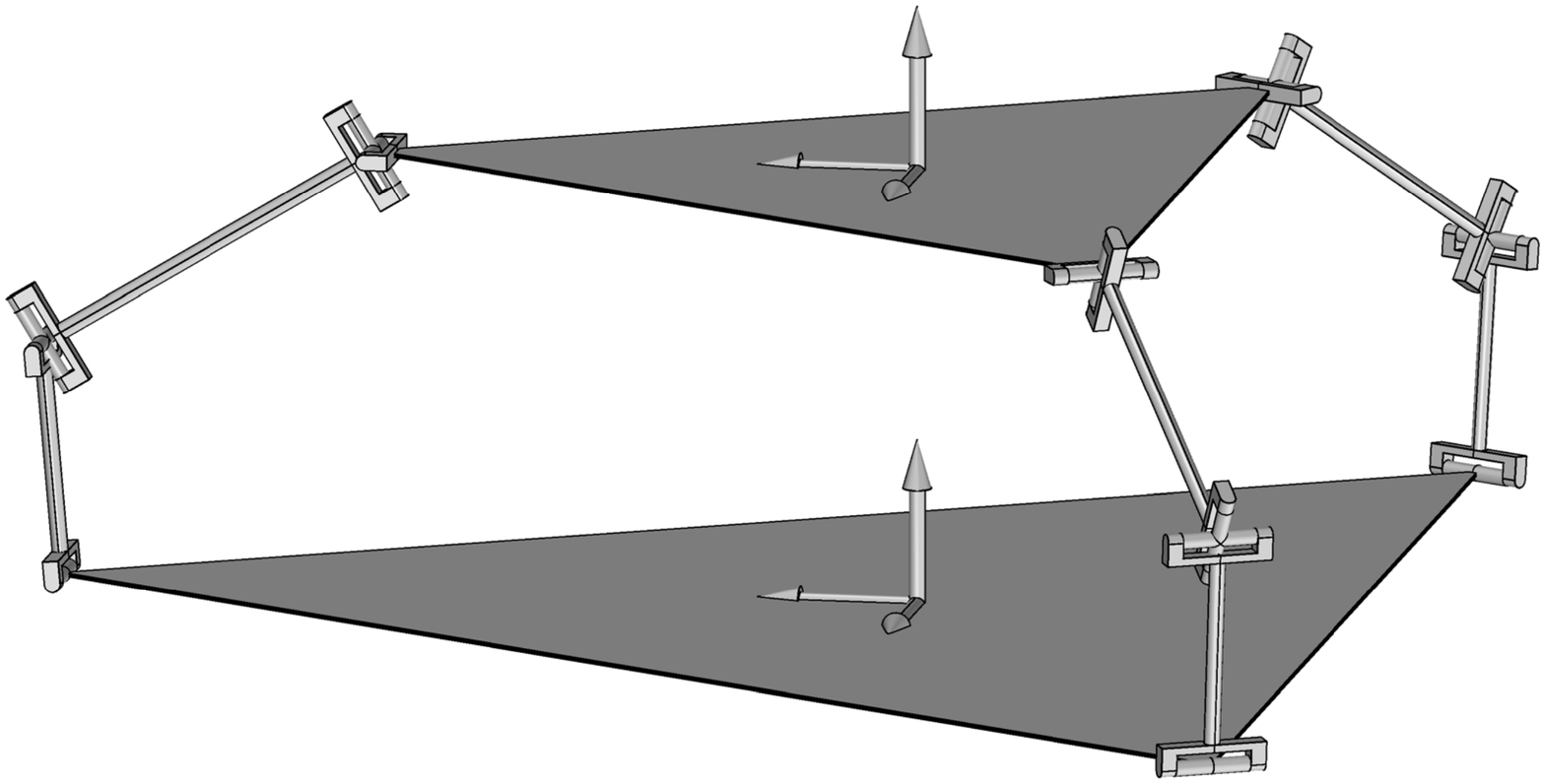}
        \put(188,146){$\Sigma_1$}
        \put(188,63){$\Sigma_0$}
        \put(28,96){$a_1$}
        \put(50,128){$a_3$}
        \put(195,180){$z_1$}
        \put(150,155){$x_1$}
        \put(175,145){$y_1$}
        \put(150,78){$x_0$}
        \put(173,65){$y_0$}
        \put(192,100){$z_0$}
    \end{overpic}\\
    \caption{3-\underline{R}UU parallel manipulator}\label{fig:3ruu}
\end{figure}

As there is a detailed explanation of the 3-\underline{R}UU parallel manipulator's
architecture in Stigger \emph{et al.}~\cite{Stigger2018} its architecture is only
briefly explained. The manipulator consists of three identical
\underline{R}UU-limbs (Fig.~\ref{fig:ruulimb}).
Each of them comprises three joints, an actuated revolute (R) and two passive
universal (U) joints. Each vertex of the equilateral base triangle is connected
via an \underline{R}UU limb with a vertex of the equilateral moving platform
triangle. The first and the last axis in each limb are tangent to the
circum-circles of base and moving triangular platform, respectively
(Fig.~\ref{fig:3ruu}). The geometric design of each limb is described with
Denavit-Hartenberg (DH) parameters~\cite{denavit-1955a}.

\subsection{Constraint Equations}\label{sec:lia}

For the workspace and singularity analysis algebraic constraint equations of the
3-\underline{R}UU chains are convenient. These equations have been derived with different
methods in~\cite{Stigger2018}. Therefore, the lengthy derivation of the
constraint equations is omitted here and the reader is referred to this paper.
The constraint equations describing a canonical \underline{R}UU limb in
Euclidean space with respect to proper coordinate systems $\Sigma_0$, $\Sigma_1$
(Fig.~\ref{fig:ruulimb}) are

\begin{align}\label{eq:idealJ1}
f_1 : & \left( \left( x_{{0}}x_{{1}}-x_{{2}}x_{{3}} \right) ({t}^{2}-1)- \left( 2 x_{{0}}x_{{2}}+2 x_{{1}}x_{{3}} \right) t \right) a_{{1}} %\nonumber \\
+2  \left( {t}^{2}+1 \right)  \left( x_{{0}}y_{{0}}+x_{{3}}y_{{3}} \right)=0, \notag\\
f_2 : & - \left( {x_{{0}}}^{2}+{x_{{1}}}^{2}+{x_{{2}}}^{2}+{x_{{3}}}^{2} \right)  \left( {t}^{2}+1 \right) {a_{{1}}}^{2}
+ \left( 4 \left(  x_{{0}}y_{{1}}- x_{{1}}y_{{0}}+ x_{{2}}y_{{3}}- x_{{3}}y_{{2}} \right) {t}^{2} \right. \nonumber \\
& + 8\left( - x_{{0}}y_{{2}}+ x_{{1}}y_{{3}}+ x_{{2}}y_{{0}}- x_{{3}}y_{{1}} \right) t
                                                                                                                               \left. -4 \left(  x_{{0}}y_{{1}}- x_{{1}}y_{{0}}+ x_{{2}}y_{{3}}- x_{{3}}y_{{2}} \right)\right) a_{{1}}\nonumber \\
& + \left(  \left( {x_{{0}}}^{2}+ {x_{{1}}}^{2}+ {x_{{2}}}^{2}+ {x_{{3}}}^{2} \right) {a_{{3}}}^{2}
-\,4 \left( {y_{{0}}}^{2}+{y_{{1}}}^{2}+{y_{{2}}}^{2}+{y_{{3}}}^{2} \right) \right)  \left( {t}^{2}+1 \right)=0.
\end{align}
The parameters $a_1$ and $a_3$ denote the leg lengths. Further, $x_i$,
$y_i$ for $i=0$, $1$, $2$, $3$ are the coordinates (Study parameters)
in a seven-dimensional projective space $\PS$, also called kinematic
image space or Study space. They determine the possible poses of the
moving coordinate system $\Sigma_1$ with respect to the base system
$\Sigma_0$. The parameter $t$ is the algebraic value, using tangent
half-angle substitution, of the input angle of the first revolute
joint. From the canonical constraint equations in
Eq.~\eqref{eq:idealJ1}, the general ones are found by applying
transformations that move the limbs from the origin of the base system
to their appropriate positions on the manipulator. Mathematically,
they are obtained by performing transformations of the $x_i$, $y_i$
coordinates in $\PS$ that are applied to the set of canonical
constraint equations. It is important to note that these
transformations do not change the degree of the constraint
equations~\cite{HuPfuSchBru}. For each limb the appropriate
transformations are applied to $f_1$ and $f_2$ which move each of the
limbs to the respective corners of the base triangle (Fig.~\ref{fig:3ruu}). Additionally, the input parameter $t$ is
substituted with $t_1$, $t_2$ and $t_3$, respectively. The radii of
the base and moving platforms circum-circles are denoted by $r_0$ and
$r_1$, respectively. In total this results in six general constraint
equations $g_1, \ldots, g_6$. Together with the equation of the Study
quadric $g_7: x_0y_0+x_1y_1+x_2y_2+x_3y_3=0$ and a normalization
condition $g_8 : x_0^2 + x_1^2 + x_2^2 + x_3^2 = 1$ these equations
yield a complete kinematic description of the manipulator, involving
the three input parameters $t_1$, $t_2$ and $t_3$. Therefore the set
${\cal W}=\{g_1,\ldots,g_8\}$ forms a three parameter system of
constraint equations in the Study parameters which define the three
dimensional constraint variety in $\mathbb{R}^{11}$.

\section{Workspace Description}\label{sec:2}

For path planning it would be convenient to have a system of equations without
the input parameters describing the workspace only in image space coordinates.
Inspection of the constraint equations shows that they are only pairwise
dependent on the same input parameter: $g_1$ and $g_2$ for example depend only
on $t_1$. Eliminating $t_1$ via computing the resultant of $g_1$ and $g_2$
yields an equation depending on the Study parameters only. Using the same
procedure on $g_3$, $g_4$ thereby eliminating $t_2$ as well as on $g_5$, $g_6$
eliminating $t_3$, three equations $g_{12}$, $g_{34}$, $g_{56}$ are obtained.
These three resulting equations are independent of the input parameters and
yield together with $g_7$ and $g_8$ a suitable description of the workspace for
path planning. As the computation is straightforward from the set of equations
\eqref{eq:idealJ1} and due to their length the three equations $g_{12}$,
$g_{34}$ and $g_{56}$ are not shown. But it is easy to show that the whole
translational three-space given by
\begin{equation}\label{transl:cond}
  {  x_0}=1,~{  x_1}={  x_2}={  x_3}=y_0= 0
\end{equation}
lies in the workspace, a fact that confirms once more that the manipulator has a
translational operation mode denoted by $O_1$, although it is not the only mode.
In order to determine all operation modes, one needs to analyze the irreducible
components of the workspace variety. Unfortunately the equations are too
complicated to perform a primary decomposition. However, the open source
software Bertini$^{\text{TM}}$~\cite{bertini2013}, a numerical algebraic
geometry software for solving polynomial systems, can be used to gain an idea
about the potential number of the components as well as their particular
dimension and degree. As Bertini$^{\text{TM}}$ is a numerical software, real
values for the leg lengths and the radii of the circum-circles of the base and
moving triangle have to be set. Running Bertini$^{\text{TM}}$ for different
designs resulted in equivalent outputs. In the following, the design
\begin{equation}
  \label{eq:pars}
  a_1 = 3, ~ a_3 = 5, ~ r_0 = 11, ~ r_1 = 7,
\end{equation}
is used as an example, Fig.~\ref{fig:3ruu}. According to~\cite{bertini2013},
results of this software hold true with very high probability depending on the
numerical error. With the equations $g_{12}$, $g_{34}$, $g_{56}$, $g_7$, $g_8$
as input Bertini$^{\text{TM}}$ classifies three components, each of dimension
three. Besides the translational three-space \eqref{transl:cond} there is
another linear component. Although Bertini$^{\text{TM}}$ does not provide an
algebraic description of the detected components, the knowledge about their
dimension and degree restricts the set of potential candidates and one finds the
second linear component to be given by
\begin{equation}\label{twist-transl:cond}
  {  x_3}=1,~{  x_0}={  x_1}={  x_2}=y_3= 0.
\end{equation}
This is again a three-space in $\PS$ denoted by $O_2$. It describes a ``twisted
translational operation mode'', that is a translational mode where the moving
platform is rotated by a half turn about the $z$-axis with respect to the base
platform. Again, it is easy to verify, that this result is independent of the
chosen design parameters. According to the output of Bertini$^{\text{TM}}$,
there exists a third component, denoted by $O_3$, of degree $944$ which suggests
that it is a rather difficult task to compute generators of a defining ideal as
well as determining whether this component is irreducible.
\section{Singularity Analysis of the Translational Operation Mode}\label{sec:3}

Singularities in parallel mechanisms can occur either because one limb or the
platform itself is in a singular position. Furthermore, both can take place at
the same time meaning that a limb and the platform are in singular positions. In
Section~\ref{sec:2} operation modes of the 3-\underline{R}UU are identified.
Below, a detailed analysis and characterization of the input and output
singularities of the translational operation mode $O_1$ will be given. In
contrast to constraint singularities, where the tangent space gains dimensions,
these singularities are points at which the tangent space contains elements
independent of the input or Study parameters, respectively. The tangent space at
a point is the kernel of the Jacobian $\mathbf{J}=[\mathbf{J_o},\mathbf{J_i}]$
evaluated at this point. Here $\mathbf{J_o}$ is the derivative of the constraint
equations with respect to the Study parameters and $\mathbf{J_i}$ is the
derivative with respect to the input parameters. The subscript $o$ refers to
\textit{output} and $i$ refers to \textit{input}. In the following input and
output singularities for the translational mode will be discussed separately in
joint space as well as in the kinematic image space. Provided the conditions
\eqref{transl:cond} hold, $g_1$, $g_3$, $g_5$, $g_7$ and $g_8$ become trivial
(meaning here that the left hand sides of these equations vanish, yielding
$0=0$). Hence the system $\cal W$ reduces to a system of three equations
\begin{align*}
    g_2 : &\left({{ a_1}}^{2}-{{
          a_3}}^{2}+(r_0-r_1)^2+4(y_1^2+y_2^2+y_3^2-a_1y_3+r_0y_1-r_1y_1)
    \right) {{ t_1}}^{2}\\
    &+ 4a_1\left( { r_0}-{ r_1}+2{ y_1} \right) { t_1}\\
    &+(r_0-r_1)^2+4(y_1^2+y_2^2+y_3^2+a_1y_3+r_0y_1-r_1y_1)+a_1^2-a_3^2 = 0,\\
    g_4 : &\left(2(r_0-r_1)\left( \sqrt{3}y_2-y_1 \right)
      +(r_0-r_1)^2+4(y_1^2+y_2^2+y_3^2-a_1y_3)+a_1^2-a_3^2\right)
    {{ t_2}}^{2}\\
    &+4a_1\left( \sqrt{3}y_2+r_0-r_1-y_1 \right){ t_2}\\
    &
    +2(r_0-r_1)(\sqrt{3}y_2-y_1)+(r_0-r_1)^2+4(y_1^2+y_2^2+y_3^2+a_1y_3)+a_1^2-a_3^2 = 0,
\end{align*}
    \begin{align*}
    g_6 : &\left(-2(r_0-r_1)(\sqrt{3}y_2+y_1)+(r_0-r_1)^2+4(y_1^2+y_2^2+y_3^2-a_1y_3)+a_1^2-a_3^2     \right) {{ t_3}}^{2}\\
    &+4a_1\left(-\sqrt{3}y_2 +r_0-r_1-y_1\right)t_3\\
    & -2(r_0-r_1)(\sqrt{3}y_2+y_1)+(r_0-r_1)^2+4(y_1^2+y_2^2+y_3^2+a_1y_3)+a_1^2-a_3^2 = 0.
\end{align*}
This system of equations is denoted by $\cal{W_T}$ and describes the
translational operation mode for arbitrary design. Note that in order to cover
all singular points on the variety defined by $\cal{W_T}$, one still has to
compute the Jacobian of the system $\cal{W}$ before substituting the conditions
for the translational operation mode in Eq.~\eqref{transl:cond} into
$\mathbf{J}$.

\subsection{Input Singularities}

\begin{figure}[bt]
  \centering
  \includegraphics[width=0.5\linewidth]{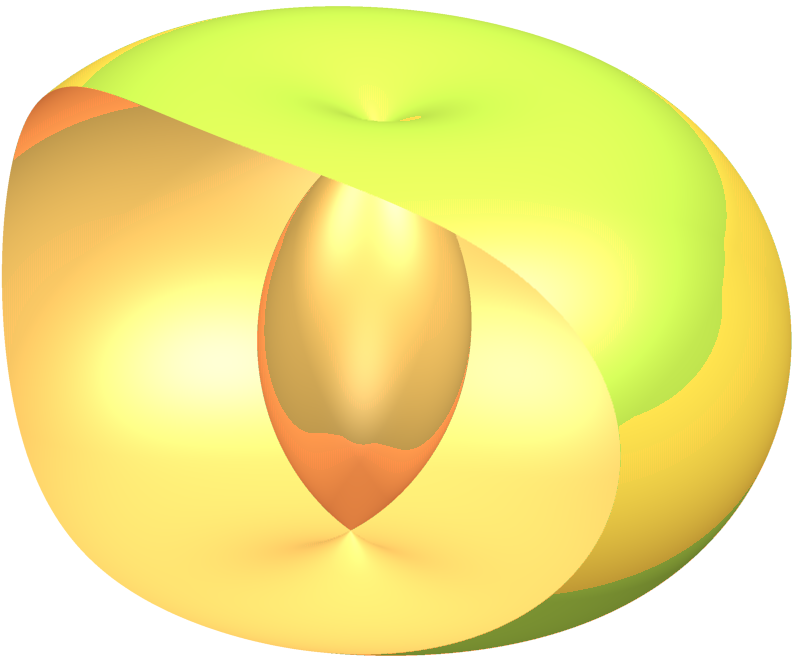}\\
  \caption{Spindle torus as input singularity surface for one limb in Study space}\label{spin:torus}
\end{figure}

Following Gosselin and Angeles~\cite{Gosselin1990}, input singularities occur
when $\mathbf{J_i}$ is rank deficient. As there are two general constraints per
limb plus the Study condition $g_7$ and the normalization condition $g_8$, the
system $\cal W$ consists of eight equations. On the other hand there are only
three input variables. The resulting $8\times3$ Jacobian matrix is obviously not
square. The algebraic condition for rank deficiency is that all $3 \times 3$
sub-determinants vanish. When the conditions for the translational mode
\eqref{transl:cond} are substituted into the system of sub-determinants, all
equations but one become trivial. The only remaining one factors into $p_1\cdot
p_2\cdot p_3 = 0$ with
\begin{equation}\label{rem:det}
  \begin{aligned}
        p_1 : &\left({{ a_1}}^{2}-{{
          a_3}}^{2}+(r_0-r_1)^2+4(y_1^2+y_2^2+y_3^2-a_1y_3+r_0y_1-r_1y_1)
    \right) {{ t_1}}\\
    &+ 2a_1\left( { r_0}-{ r_1}+2{ y_1} \right),\\
    p_2 : &\left(2(r_0-r_1)\left( \sqrt{3}y_2-y_1 \right)
      +(r_0-r_1)^2+4(y_1^2+y_2^2+y_3^2-a_1y_3)+a_1^2-a_3^2\right)
    {{ t_2}}\\
    &+2a_1\left( \sqrt{3}y_2+r_0-r_1-y_1 \right),\\
    p_3 : &\left(-2(r_0-r_1)(\sqrt{3}y_2+y_1)+(r_0-r_1)^2+4(y_1^2+y_2^2+y_3^2-a_1y_3)+a_1^2-a_3^2     \right) {{ t_3}}\\
    &+2a_1\left(-\sqrt{3}y_2 +r_0-r_1-y_1\right).
  \end{aligned}
\end{equation}
Each factor is related to one limb of the mechanism. Since
it is sufficient that one of the three factors vanishes in order to satisfy the
singularity condition only the first of the factors is discussed here in detail.
The other ones can be treated analogously. As $p_1$ in Eq.~\eqref{rem:det} is
linear in $t_1$, it has the unique root at
\begin{align}\label{root:p1}
  t_1 = \frac {2{ a_1}  \left( { r_1}-{ r_0}-2 { y_1} \right) }
  {\left(
  {{ a_1}}^{2}-{{a_3}}^{2}+(r_0-r_1)^2+4(y_1^2+y_2^2+y_3^2-a_1y_3+r_0y_1-r_1y_1)
  \right)},
\end{align}
provided that the leading coefficient of $p_1$ with respect to $t_1$, i.e. the
denominator in \eqref{root:p1}, does not vanish. Substituting $t_1$ from
Eq.~\eqref{root:p1} into the system $\cal W_T$, its first equation simplifies
to
\begin{equation}
  \label{tor:1}
  \begin{aligned}
    0= &~{{ a_1}}^{4}-2 {{ a_3}}^{2}{{ a_1}}^{2}-2 {{ r_0}}^{2}{{ a_1}}^{2}+4 { r_0}
    { r_1} {{ a_1}}^{2}-8 { r_0} {{ a_1}} ^{2}{ y_1}-2 {{ r_1}}^{2}{{
        a_1}}^{2}+8 { r_1} {{ a_1}}^{2}{ y_1}\\
    &-8 {{ a_1}}^{2}{{ y_1}}^{2}+8 {{ a_1}}^{2}{{ y_2}} ^{2}-8 {{ a_1}}^{2}{{
        y_3}}^{2}+{{ a_3}}^{4}-2 {{ a_3}}^{2}{ { r_0}}^{2}+4 {{ a_3}}^{2}{ r_0}
    { r_1}-8 {{ a_3}}^{2}{r_0} { y_1}\\
    &-2 {{ a_3}}^{2}{{ r_1}}^{2}+8 {{ a_3}}^{2}{ r_1} { y_1}-8 {{ a_3}}^{2}{{
        y_1}}^{2}-8 {{ a_3}}^{2}{{ y_2}}^{2}-8 {{ a_3}}^{2}{{ y_3}}^{2}+{{
        r_0}}^{4}-4 {{ r_0}}^{3}{ r_1}\\
    &+8 {{ r_0}}^{3}{ y_1}+6 {{ r_0}}^{2}{{r_1}} ^{2}-24 {{ r_0}}^{2}{ r_1} {
      y_1}+24 {{ r_0}}^{2}{{ y_1}} ^{2}+8 {{ r_0}}^{2}{{ y_2}}^{2}+8 {{
        r_0}}^{2}{{ y_3}}^{2}-4 { r_0} {{ r_1}}^{3}\\
    &+24 { r_0} {{ r_1}}^{2}{ y_1}-48 { r_0} { r_1} {{ y_1}}^{2}-16 { r_0} {
      r_1} {{ y_2}}^{ 2}-16 { r_0} { r_1} {{ y_3}}^{2}+32 { r_0} {{ y_1}}^{3}
    +32 { r_0} { y_1} {{ y_2}}^{2}\\
    &+32 { r_0} { y_1} {{ y_3}}^{2}+{{ r_1}}^{4}-8 {{ r_1}}^{3}{ y_1}+24 {{
        r_1}}^{2}{{ y_1}}^{2}+8 {{ r_1}}^{2}{{ y_2}}^{2}+8 {{ r_1}}^{2}{{ y_3
      }}^{2}-32 { r_1} {{ y_1}}^{3}\\
    &-32 { r_1} { y_1} {{ y_2}} ^{2}-32 { r_1} { y_1} {{ y_3}}^{2}+16 {{
        y_1}}^{4}+32 {{ y_1}}^{2}{{ y_2}}^{2}+32 {{
        y_1}}^{2}{{ y_3}}^{2}+16 {{ y_2}}^{4}\\
    &+32 {{ y_2}}^{2}{{ y_3}}^{2}+16 {{ y_3}}^{4}
  \end{aligned}
\end{equation}
Since the other two equations of $\cal W_T$ involve only $t_2$ and $t_3$,
respectively, one always finds values for these input parameters such that these
equations hold for $y_1$, $y_2$, $y_3$ fulfilling Eq.~\eqref{tor:1}. It remains
to discuss the case when the leading coefficient of $p_1$ with respect to $t_1$
is zero, i.e.
\begin{align}\label{leadcoeff}
{{ a_1}}^{2}-4 { a_1} { y_3}-{{ a_3}}^{2}+{{ r_0}}^{2}-2 
{ r_0} { r_1}+4 { r_0} { y_1}+{{ r_1}}^{2}-4 { r_1} {
 y_1}+4 {{ y_1}}^{2}+4 {{ y_2}}^{2}+4 {{ y_3}}^{2}=0
\end{align}
In this case, the constant term of $p_1$ needs to be zero as well, that is
$y_1=(r_1-r_0)/2$. Substituting this condition into $\cal W_T$ and simplifying
with \eqref{leadcoeff}, one obtains the further condition $y_3=0$. Moreover one
gets $y_2=\pm \sqrt{(a_3^2-a_1^2)/4}$ via substituting $y_1 = (r_1-r_0)/2$ and
$y_3 = 0$ into Eq.~\eqref{leadcoeff}. However the two points
$\left((r_1-r_0)/2,\pm \sqrt{(a_3^2-a_1^2)/4},0\right)$ already lie on the
surface defined by Eq.~\eqref{tor:1}, thus this equation together with the
system $\cal{W_T}$ yields the singularity surface in the kinematic image space
related to the first limb. It has the following interpretation: when a pose of
the manipulator in the translational operation mode is chosen such that
Eq.~\eqref{tor:1} is fulfilled, then the first limb is either stretched out or
folded. The surface can be visualized in the three dimensional Study subspace
spanned by $y_1$, $y_2$, $y_3$ representing all translations.
Fig.~\ref{spin:torus} shows the input singularity surface for the first limb of
the 3-\underline{R}UU PM with the design parameters in Eq.~\eqref{eq:pars}. An
easy computation shows that it is a torus.
The other two factors in Eq.~\eqref{rem:det} yield also tori which are rotated
by $\pm 120^{\circ}$ about the $y_3$-axis. All three surfaces have two points in
common. These points correspond to poses where all three limbs are singular.
Intersection curves of a pair of tori correspond to poses where two limbs are
singular.

\begin{figure}[ht!]
  \centering
  \includegraphics[width=0.4\linewidth]{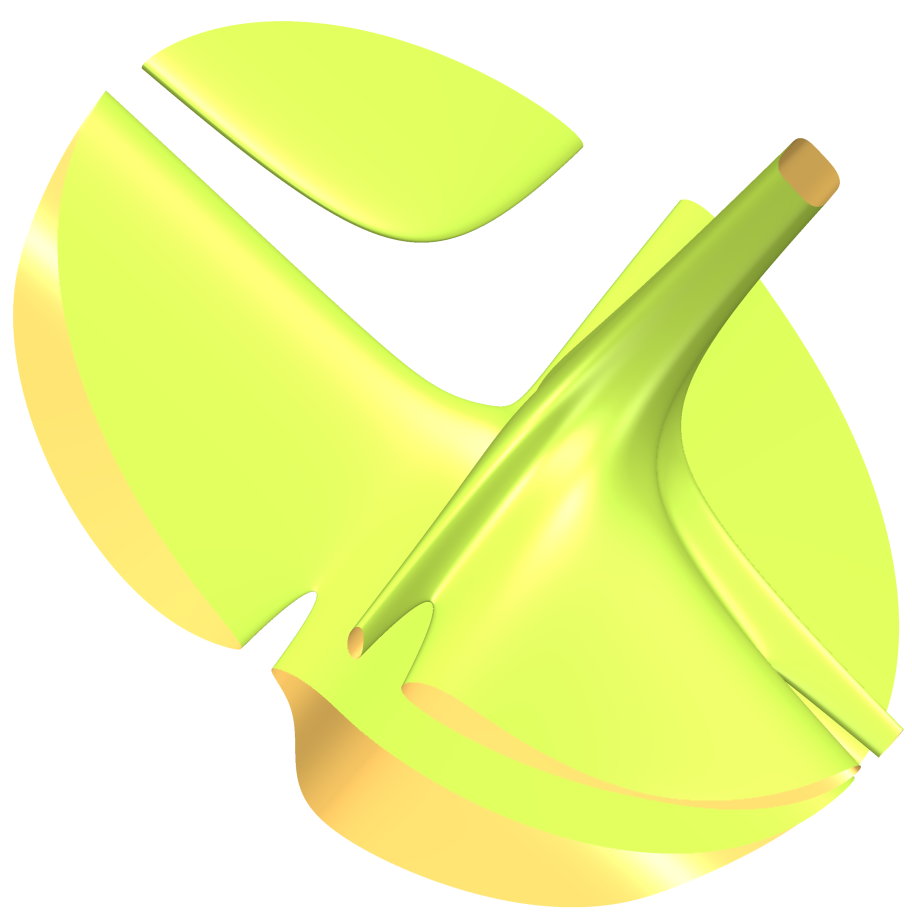}\\
  \caption{Input singularity surface for one limb in the joint space}\label{vier:klee}
\end{figure}

For practical control purposes it would be convenient to have a
representation of the input singularities in joint space, as to avoid
these inputs during operation. In order to simplify computations and
display the results the design parameters in Eq.~\eqref{eq:pars} are
set. However, the procedure below can be performed for any design. A
condition for the first limb to be in a singular position is found by
computing a Groebner basis of the ideal generated by the system
$\cal W_T$ and $p_1$ in Eq.~\eqref{rem:det} with respect to a proper
order, such that the first polynomial in the base is independent of
the Study parameters. In the following, the product order on
$\mathbb{R}[y_1,y_2,y_3]\times \mathbb{R}[t_1,t_2,t_3]$, where each of
these rings is equipped with the graded reverse lexicographic order is
used. The first polynomial in the base is of degree $12$ and depends
on the input parameters only. It defines a variety in joint space
containing all inputs, such that at least one solution of the direct
kinematics yields a position in which the first limb of the PM is
singular. This polynomial reads
\begin{small}
\begin{gather*}
  32 {{ t_1}}^{4}{{ t_2}}^{4}{{ t_3}}^{4}+96 {{ t_1}}^{4}{{ t_2}}^{4}{{
      t_3}}^{3}+96 {{ t_1}}^{4}{{ t_2}}^{3}{{ t_3}}^ {4}+96 {{ t_1}}^{3}{{
      t_2}}^{4}{{ t_3}}^{4}-16 {{ t_1}}^{4}{ { t_2}}^{4}{{ t_3}}^{2}+240 {{
      t_1}}^{4}{{ t_2}}^{3}{{ t_3}
  }^{3}\\
  +160 {{ t_1}}^{4}{{ t_2}}^{2}{{ t_3}}^{4}+240 {{ t_1}}^ {3}{{ t_2}}^{4}{{
      t_3}}^{3}+288 {{ t_1}}^{3}{{ t_2}}^{3}{{ t_3}}^{4}+160 {{ t_1}}^{2}{{
      t_2}}^{4}{{ t_3}}^{4}-32 {{ t_1}}^{4}{{ t_2}}^{4}{ t_3}+44 {{ t_1}}^{4}{{
      t_2}}^{3}{{
      t_3}}^{2}\\
  +272 {{ t_1}}^{4}{{ t_2}}^{2}{{ t_3}}^{3}+132 {{ t_1}}^{4}{ t_2} {{
      t_3}}^{4}+44 {{ t_1}}^{3}{{ t_2}}^{4}{ { t_3}}^{2}+576 {{ t_1}}^{3}{{
      t_2}}^{3}{{ t_3}}^{3}+480 {{ t_1}}^{3}{{ t_2}}^{2}{{ t_3}}^{4}+272 {{
      t_1}}^{2}{{ t_2}}
  ^{4}{{  t_3}}^{3}\\
  +480 {{ t_1}}^{2}{{ t_2}}^{3}{{ t_3}}^{4}+132 { t_1} {{ t_2}}^{4}{{
      t_3}}^{4}+48 {{ t_1}}^{4}{{ t_2}}^ {4}+16 {{ t_1}}^{4}{{ t_2}}^{3}{
    t_3}+102 {{ t_1}}^{4}{{ t_2}}^{2}{{ t_3}}^{2}+80 {{ t_1}}^{4}{ t_2} {{
      t_3}}^{3}+72
  {{  t_1}}^{4}{{  t_3}}^{4}\\
  +16 {{ t_1}}^{3}{{ t_2}}^{4}{ t_3}+ 152 {{ t_1}}^{3}{{ t_2}}^{3}{{
      t_3}}^{2}+740 {{ t_1}}^{3}{{ t_2}}^{2}{{ t_3}}^{3}+396 {{ t_1}}^{3}{ t_2}
  {{ t_3}}^{4} +102 {{ t_1}}^{2}{{ t_2}}^{4}{{ t_3}}^{2}+740 {{ t_1}}^{2}{{
      t_2}}^{3}{{  t_3}}^{3}\\
  +718 {{ t_1}}^{2}{{ t_2}}^{2}{{ t_3}} ^{4}+80 { t_1} {{ t_2}}^{4}{{
      t_3}}^{3}+396 { t_1} {{ t_2}}^{3}{{ t_3}}^{4}+72 {{ t_2}}^{4}{{
      t_3}}^{4}+156 {{ t_1} }^{4}{{ t_2}}^{3}+192 {{ t_1}}^{4}{{ t_2}}^{2}{
    t_3}-52 {{
      t_1}}^{4}{  t_2} {{  t_3}}^{2}\\
  +32 {{ t_1}}^{4}{{ t_3}}^{3}+ 156 {{ t_1}}^{3}{{ t_2}}^{4}+409 {{ t_1}}^{3}{{
      t_2}}^{2}{{ t_3}}^{2}+288 {{ t_1}}^{3}{ t_2} {{ t_3}}^{3}+216 {{ t_1
    }}^{3}{{ t_3}}^{4}+192 {{ t_1}}^{2}{{ t_2}}^{4}{ t_3}+409 {{
      t_1}}^{2}{{  t_2}}^{3}{{  t_3}}^{2}\\
  +962 {{ t_1}}^{2}{{ t_2}} ^{2}{{ t_3}}^{3}+537 {{ t_1}}^{2}{ t_2} {{
      t_3}}^{4}-52 { t_1} {{ t_2}}^{4}{{ t_3}}^{2}+288 { t_1} {{ t_2}}^{3}{{
      t_3}}^{3}+537 { t_1} {{ t_2}}^{2}{{ t_3}}^{4}+32 {{ t_2}
  }^{4}{{  t_3}}^{3}+216 {{  t_2}}^{3}{{  t_3}}^{4}\\
  +266 {{ t_1}}^ {4}{{ t_2}}^{2}+176 {{ t_1}}^{4}{ t_2} { t_3}-80 {{ t_1}}
  ^{4}{{ t_3}}^{2}+360 {{ t_1}}^{3}{{ t_2}}^{3}+348 {{ t_1}}^{ 3}{{ t_2}}^{2}{
    t_3}+8 {{ t_1}}^{3}{ t_2} {{ t_3}}^{2}+176
  {{  t_1}}^{3}{{  t_3}}^{3}\\
  +266 {{ t_1}}^{2}{{ t_2}}^{4}+348 {{ t_1}}^{2}{{ t_2}}^{3}{ t_3}+996 {{
      t_1}}^{2}{{ t_2}}^{2} {{ t_3}}^{2}+348 {{ t_1}}^{2}{ t_2} {{ t_3}}^{3}+266
  {{ t_1}}^{2}{{ t_3}}^{4}+176 { t_1} {{ t_2}}^{4}{ t_3}+8 {
    t_1} {{  t_2}}^{3}{{  t_3}}^{2}\\
  +348 { t_1} {{ t_2}}^{2}{{ t_3}}^{3}+360 { t_1} { t_2} {{ t_3}}^{4}-80 {{
      t_2}}^{4}{{ t_3}}^{2}+176 {{ t_2}}^{3}{{ t_3}}^{3}+266 {{ t_2}}^{2}{{
      t_3}}^{4}+216 {{ t_1}}^{4}{ t_2}+32 {{ t_1}}^{4}{ t_3}+
  537 {{  t_1}}^{3}{{  t_2}}^{2}\\
  +288 {{ t_1}}^{3}{ t_2} { t_3 }-52 {{ t_1}}^{3}{{ t_3}}^{2}+537 {{ t_1}}^{2}{{
      t_2}}^{3}+ 962 {{ t_1}}^{2}{{ t_2}}^{2}{ t_3}+409 {{ t_1}}^{2}{ t_2} {{
      t_3}}^{2}+192 {{ t_1}}^{2}{{ t_3}}^{3}+216 { t_1} {{
      t_2}}^{4}\\
  +288 { t_1} {{ t_2}}^{3}{ t_3}+409 { t_1} {{ t_2}}^{2}{{ t_3}}^{2}+156 { t_1}
  {{ t_3}}^{4}+32 {{ t_2} }^{4}{ t_3}-52 {{ t_2}}^{3}{{ t_3}}^{2}+192 {{
      t_2}}^{2}{{
      t_3}}^{3}+156 {  t_2} {{  t_3}}^{4}+72 {{  t_1}}^{4}\\
  +396 {{ t_1}}^{3}{ t_2}+80 {{ t_1}}^{3}{ t_3}+718 {{ t_1}}^{2}{{ t_2}}^{2}+740
  {{ t_1}}^{2}{ t_2} { t_3}+102 {{ t_1}}^{2} {{ t_3}}^{2}+396 { t_1} {{
      t_2}}^{3}+740 { t_1} {{ t_2}}
  ^{2}{  t_3}+152 {  t_1} {  t_2} {{  t_3}}^{2}\\
  +16 { t_1} {{ t_3}}^{3}+72 {{ t_2}}^{4} +80 {{ t_2}}^{3}{ t_3}+102 {{
      t_2}}^{2}{{ t_3}}^{2}+16 { t_2} {{ t_3}}^{3}+48 {{ t_3}}^{4} +132 {{
      t_1}}^{3}+480 {{ t_1}}^{2}{ t_2}+272 {{ t_1}}^{2}{
    t_3}\\
  +480 { t_1} {{ t_2}}^{2}+576 { t_1} { t_2} { t_3 }+44 { t_1} {{ t_3}}^{2} +132
  {{ t_2}}^{3}+272 {{ t_2}}^{2} { t_3}+44 { t_2} {{ t_3}}^{2}-32 {{
      t_3}}^{3}+160 {{ t_1
    }}^{2}+288 {  t_1} {  t_2}\\
  +240 { t_1} { t_3} +160 {{ t_2}} ^{2}+240 { t_2} { t_3}-16 {{ t_3}}^{2}+96 {
    t_1}+96 { t_2}+96 { t_3}+32.
\end{gather*}
\end{small}
The singularity surface in the joint space is displayed in Fig.~\ref{vier:klee}.
The singularity surfaces of the other limbs can be computed analogously, using
the factors $p_2$ or $p_3$ in Eq.~\eqref{rem:det}. They are congruent to the
surface in Fig.~\ref{vier:klee}.

\subsection{Output Singularities}\label{sec:output}

In order to compute the output singularities, one has to check if $\mathbf{J_o}$
is rank deficient. For the 3-\underline{R}UU PM there are 8 constraint
equations and also 8 Study parameters. Thus $\mathbf{J_o}$ is an $8\times8$
matrix and therefore it is rank deficient if the condition
$\det(\mathbf{J_o})=0$ holds. As the aim of this paper is to investigate the
translational operation mode the conditions \eqref{transl:cond} are substituted
into $\mathbf{J_o}$. The determinant of the resulting matrix factors into
$s_1\cdot s_2$ (equation shown in Appendix). When $s_1$ and $s_2$ are computed with general
design parameters, then these equations are rather lengthy. They are therefore
contained in the appendix.

The system $\cal W_T$ together with the condition $\det(\mathbf{J_o})=0$ yields a
system $\cal L$ of four equations which describe all output singularities in the
translational operation mode $O_1$. Note that $s_1$ and $s_2$ are computed for
arbitrary design and do not depend on the second leg length $a_3$. To derive the
output singularities in the kinematic image space, the design parameters in
Eq.~\eqref{eq:pars} are set. Again, the following procedure can be performed
with any design. One eliminates $t_1$, $t_2$ and $t_3$ successively and obtains
a single equation with four significant factors $F_1$, $F_2$, $F_3$ and $F_4$.
Thus the variety decomposes into the components $\mathbb{V}(F_1)$,
$\mathbb{V}(F_2)$, $\mathbb{V}(F_3)$ and $\mathbb{V}(F_4)$. The first two
factors read
\begin{equation}\label{tor:2}
  F_{1,2}=y_1^2+y_2^2+\left( y_3\pm i\frac{\sqrt{7}}{2}\right)^2-\frac{25}{4}.
\end{equation}
These polynomials vanish on complex spheres with centers $(0,0,\pm i\sqrt{7}/2)$
and radius $5/2$ each. The real valued points of these spheres lie on the circle
in the plane $y_3=0$, given by $y_1^2+y_2^2-8=0$ which also happens to be their
intersection. For points on this circle, the initial constraint equations $g_1$,
$g_3$ and $g_5$ are fulfilled anyway. The equations $g_2$, $g_4$ and $g_6$
factor into two factors, one of them depending on the Study parameters and the
other one on the input parameters only. Thus the constraint equations read
\begin{equation}\label{eq:circle-cond}
\begin{aligned}
  \left( {{  t_1}}^{2}+\frac{3}{2} {  t_1}+1 \right)  \left( {  y_1}+2
  \right)&=0,\\
  \left( {{  t_2}}^{2}+\frac{3}{2} {  t_2}+1 \right)  \left( -\sqrt {3}
  {  y_2}+{  y_1}-4 \right)&=0,\\
  \left( {{  t_3}}^{2}+ \frac{3}{2}{  t_3}+1 \right)  \left( \sqrt {3}{
  y_2}+{  y_1}-4 \right)&=0,\\ 
{{  y_1}}^{2}+{{  y_2}}^{2}-8&=0,\\
 y_3 &= 0.
\end{aligned}
\end{equation}
This implies a self-motion on the circle $y_1^2 + y_2^2 - 8 = 0$ in
the plane $y_3=0$, since all of its points can be reached with fixed
input parameters obtained as roots of the first factors in
Eq.~\eqref{eq:circle-cond}. On the other hand
Eq.~\eqref{eq:circle-cond} is a system of five equations in six
indeterminantes. It is easy to see that one can choose an arbitrary
value for $t_1$ and the system is still solvable, meaning that the
first limb can move freely without changing the position of the end
effector. The same holds for the other two limbs choosing arbitrary
values for $t_2$ or $t_3$, respectively. In
Section~\ref{sec:selfmotion} this self-motion is discussed for
arbitrary design parameters.
\begin{figure}[h]
  \begin{minipage}{0.5\linewidth}
    \centering
    \includegraphics[width=0.7\textwidth]{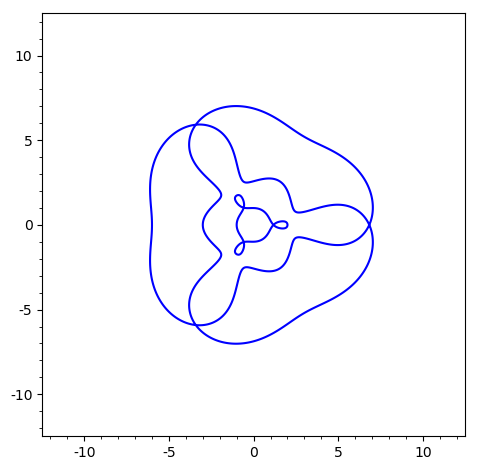}
    \caption{Intersection of $\mathbb{V}(F_3)$ at $y_3 = 0$}\label{fig:y301}
  \end{minipage}\hfill
  \begin{minipage}{0.5\linewidth}
    \centering
    \includegraphics[width=0.7\textwidth]{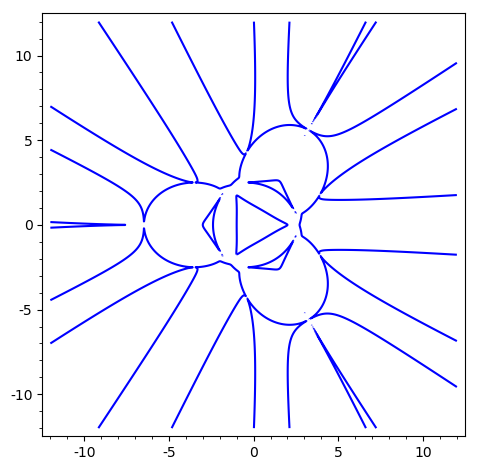}
    \caption{Intersection of $\mathbb{V}(F_4)$ at $y_3 = 0$}\label{fig:y302}
  \end{minipage}
\end{figure}

\begin{figure}[h!]
  \begin{minipage}{0.5\linewidth}
    \centering \includegraphics[width=0.9\textwidth]{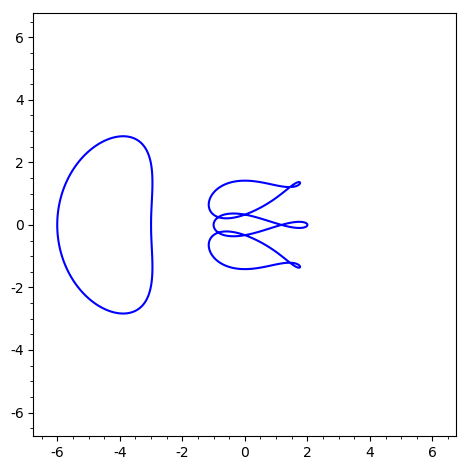}
    \caption{Intersection of $\mathbb{V}(F_3)$ at $y_2 = 0$}\label{fig:y201}
  \end{minipage}\hfill
  \begin{minipage}{0.5\linewidth}
    \centering
    \includegraphics[width=0.9\textwidth]{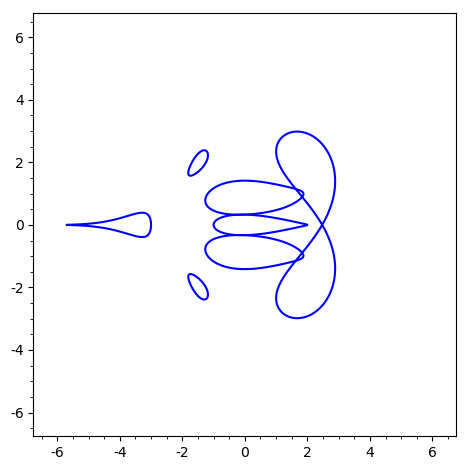}
    \caption{Intersection of $\mathbb{V}(F_4)$ at $y_2 = 0$}\label{fig:y202}
  \end{minipage}
\end{figure}

The factors $F_3$ and $F_4$ yield surfaces of degree $28$ and $48$,
respectively, which were not possible to be plotted nicely at once, but
intersections with the plane $y_3=0$ are shown in Figs.~\ref{fig:y301} and
\ref{fig:y302}, whereas intersections with $y_2=0$ are shown in
Figs.~\ref{fig:y201} and \ref{fig:y202}.

Note that if one would differentiate the system $\cal W_T$, the factor $F_4$
does not appear in the determinant of the Jacobian of this system. Since $\cal
W_T$ consists of constraint equations for the translational three-space, this
approach yields poses where the manipulator can move infinitesimally in
translational direction.

\begin{figure}[ht!]
  \begin{minipage}{0.5\linewidth}
    \centering \includegraphics[width=0.6\textwidth]{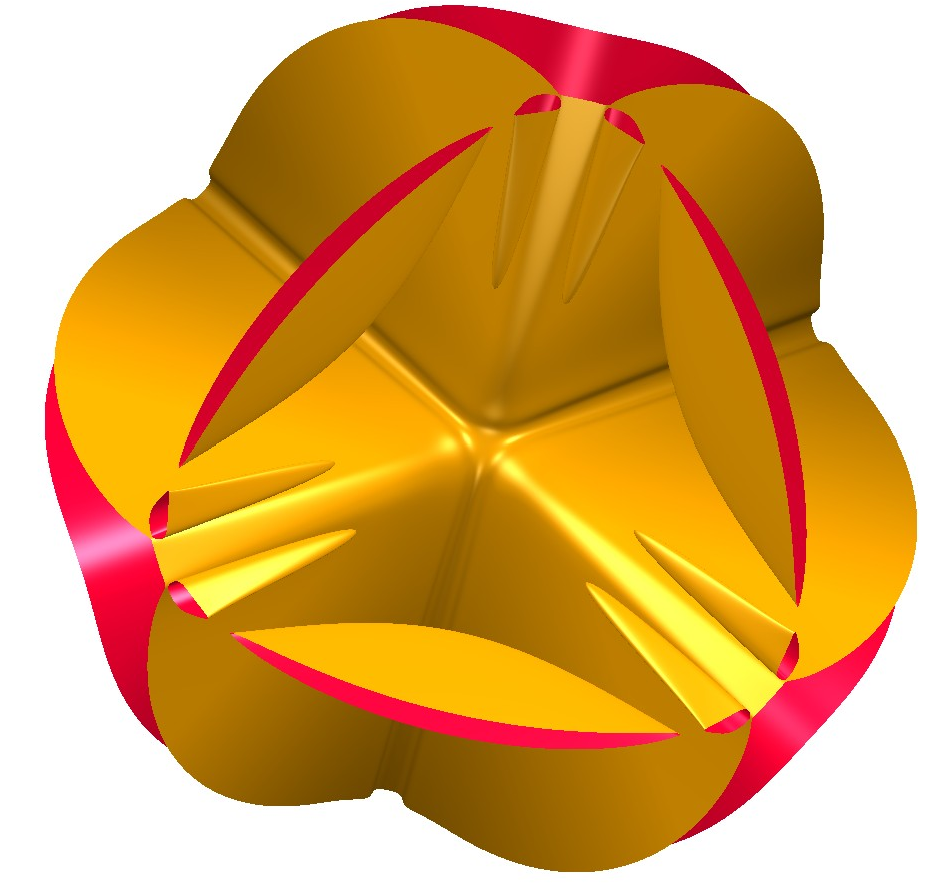}
    \caption{Simpler component of output singularities in joint space}\label{fig:mokey}
  \end{minipage}
  \begin{minipage}{0.5\linewidth}
    \centering \includegraphics[width=0.95\textwidth]{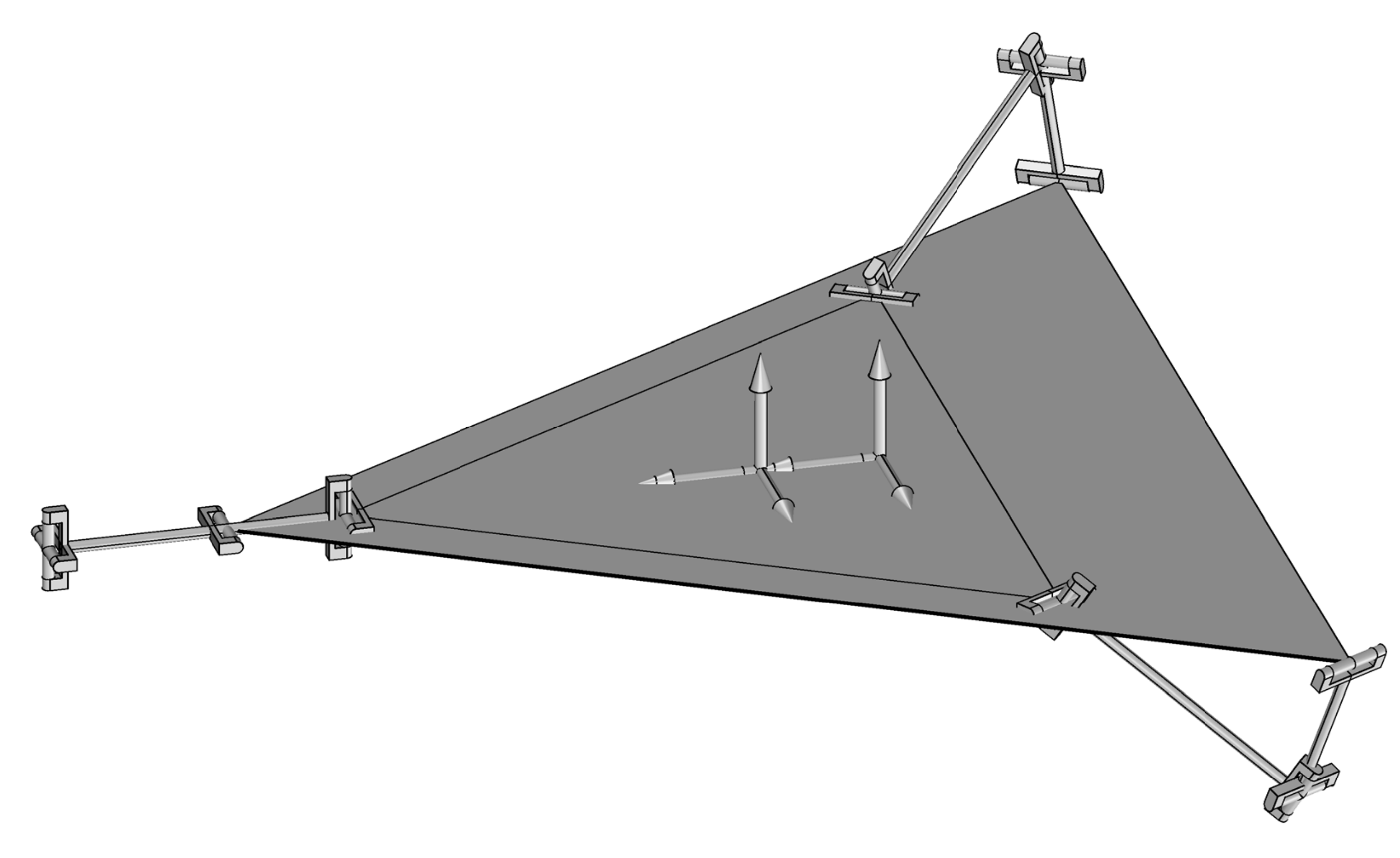}
    \caption{Simultaneous input and output singularity}\label{fig:I+Osing}
  \end{minipage}
\end{figure}

In the same way as it is done for the input singularities, it is possible to
obtain a variety in joint space containing all output singularities via
computing a Groebner basis of the system $\cal L$ that contains a polynomial
depending on the input parameters only. Due to complexity reasons, symbolic
computation was only possible for the simpler factor of the determinant of
$\mathbf{J_o}$. This yields a polynomial of degree 12 in joint space, namely
\begin{small} 
  \begin{gather*}
    144 {{ t_1}}^{4}{{ t_2}}^{4}{{ t_3}}^{4}+32 {{ t_1}}^{4}{{ t_2}}^{4}{{
        t_3}}^{3}+32 {{ t_1}}^{4}{{ t_2}}^{3}{{ t_3}}^ {4}+32 {{ t_1}}^{3}{{
        t_2}}^{4}{{ t_3}}^{4} -272 {{ t_1}}^{4} {{ t_2}}^{4}{{ t_3}}^{2} -520 {{
        t_1}}^{4}{{ t_2}}^{3}{{ t_3      }}^{3}\\
    -272 {{ t_1}}^{4}{{ t_2}}^{2}{{ t_3}}^{4}-520 {{ t_1}} ^{3}{{ t_2}}^{4}{{
        t_3}}^{3}-520 {{ t_1}}^{3}{{ t_2}}^{3}{{ t_3}}^{4}-272 {{ t_1}}^{2}{{
        t_2}}^{4}{{ t_3}}^{4}-544 {{ t_1}}^{4}{{ t_2}}^{4}{ t_3}-1352 {{
        t_1}}^{4}{{t_2}}^{3}{ { t_3}}^{2}\\
    -1352 {{ t_1}}^{4}{{ t_2}}^{2}{{ t_3}}^{3} -544 {{ t_1}}^{4}{ t_2} {{
        t_3}}^{4}-1352 {{ t_1}}^{3}{{ t_2}}^{4 }{{ t_3}}^{2}-1656 {{ t_1}}^{3}{{
        t_2}}^{3}{{ t_3}}^{3}-1352 {{ t_1}}^{3}{{ t_2}}^{2}{{ t_3}}^{4}-1352 {{
        t_1}}^{2}{{ t_2}}^{4}{{ t_3}}^{3}\\
    -1352 {{ t_1}}^{2}{{ t_2}}^{3}{{ t_3} }^{4}-544 { t_1} {{ t_2}}^{4}{{
        t_3}}^{4}+16 {{ t_1}}^{4}{{ t_2}}^{4}-1528 {{ t_1}}^{4}{{ t_2}}^{3}{
      t_3}-3183 {{ t_1 }}^{4}{{ t_2}}^{2}{{ t_3}}^{2}-1528 {{ t_1}}^{4}{ t_2} {{
        t_3}}^{3}+16 {{ t_1}}^{4}{{ t_3}}^{4}\\
    -1528 {{ t_1}}^{3}{{ t_2}}^{4}{ t_3}-3182 {{ t_1}}^{3}{{ t_2}}^{3}{{
        t_3}}^{2}- 3182 {{ t_1}}^{3}{{ t_2}}^{2}{{ t_3}}^{3}-1528 {{ t_1}}^{3}{
      t_2} {{ t_3}}^{4}-3183 {{ t_1}}^{2}{{ t_2}}^{4}{{ t_3}}^{ 2}-3182 {{
        t_1}}^{2}{{ t_2}}^{3}{{ t_3}}^{3}\\
    -3183 {{ t_1}}^{ 2}{{ t_2}}^{2}{{ t_3}}^{4}-1528 { t_1} {{ t_2}}^{4}{{ t_3}
    }^{3}-1528 { t_1} {{ t_2}}^{3}{{ t_3}}^{4}+16 {{ t_2}}^{4}{ { t_3}}^{4}-448
    {{ t_1}}^{4}{{ t_2}}^{3}-2576 {{ t_1}}^{4}{{ t_2}}^{2}{ t_3}\\
    -2576 {{t_1}}^{4}{ t_2} {{ t_3}}^{2} -448 {{ t_1}}^{4}{{ t_3}}^{3}-448 {{
        t_1}}^{3}{{ t_2}}^{4}-2520 {{ t_1}}^{3}{{ t_2}}^{3}{ t_3}-6880 {{
        t_1}}^{3}{{ t_2}}^ {2}{{ t_3}}^{2}-2520 {{ t_1}}^{3}{ t_2} {{
        t_3}}^{3}\\
    -448 {{ t_1}}^{3}{{ t_3}}^{4}-2576 {{ t_1}}^{2}{{ t_2}}^{4}{ t_3}- 6880 {{
        t_1}}^{2}{{ t_2}}^{3}{{ t_3}}^{2}-6880 {{ t_1}}^{2}{ { t_2}}^{2}{{
        t_3}}^{3}-2576 {{ t_1}}^{2}{ t_2} {{ t_3}}^{ 4}-2576 { t_1}
    {{ t_2}}^{4}{{ t_3}}^{2}\\
    -2520 { t_1} {{ t_2}}^{3}{{ t_3}}^{3}-2576 { t_1} {{ t_2}}^{2}{{ t_3}}^{4}-
    448 {{ t_2}}^{4}{{ t_3}}^{3}-448 {{ t_2}}^{3}{{ t_3}}^{4}- 928 {{
        t_1}}^{4}{{ t_2}}^{2}-1480 {{ t_1}}^{4}{ t_2} { t_3}-928 {{ t_1}}^{4}{{
        t_3}}^{2}\\
    -1480 {{ t_1}}^{3}{{ t_2}}^{ 3}-4562 {{ t_1}}^{3}{{ t_2}}^{2}{ t_3}-4562 {{
        t_1}}^{3}{ t_2} {{ t_3}}^{2}-1480 {{ t_1}}^{3}{{ t_3}}^{3}-928 {{
        t_1}}^{2}{{ t_2}}^{4}-4562 {{ t_1}}^{2}{{ t_2}}^{3}{ t_3}\\
    - 12486 {{t_1}}^{2}{{ t_2}}^{2}{{ t_3}}^{2} -4562 {{ t_1}}^{2} { t_2} {{
        t_3}}^{3}-928 {{ t_1}}^{2}{{ t_3}}^{4}-1480 { t_1} {{ t_2}}^{4}{
      t_3}-4562 { t_1} {{ t_2}}^{3}{{ t_3}}^{ 2}-4562 { t_1} {{ t_2}}^{2}{{
        t_3}}^{3}\\
    -1480 { t_1} { t_2} {{ t_3}}^{4}-928 {{ t_2}}^{4}{{ t_3}}^{2}-1480 {{ t_2}}
    ^{3}{{ t_3}}^{3}-928 {{ t_2}}^{2}{{ t_3}}^{4}-448 {{ t_1}}^{ 4}{ t_2}-448 {{
        t_1}}^{4}{ t_3}-2576 {{ t_1}}^{3}{{ t_2}}^ {2}\\
    -2520 {{ t_1}}^{3}{ t_2} {t_3} -2576 {{ t_1}}^{3}{{ t_3}}^{2}-2576 {{
        t_1}}^{2}{{ t_2}}^{3}-6880 {{ t_1}}^{2}{{ t_2}}^{2}{ t_3}-6880 {{
        t_1}}^{2}{ t_2} {{ t_3}}^{2}-2576 {{ t_1}}^{2}{{ t_3}}^{3}-448 { t_1} {{
        t_2}}^{4}\\
    -2520 { t_1 } {{ t_2}}^{3}{ t_3}-6880 { t_1} {{ t_2}}^{2}{{ t_3}}^{2} -2520
    { t_1} { t_2} {{ t_3}}^{3}-448 { t_1} {{ t_3}}^{4 }-448 {{ t_2}}^{4}{
      t_3}-2576 {{ t_2}}^{3}{{ t_3}}^{2}-2576 {{ t_2}}^{2}{{ t_3}}^{3}\\
    -448 {t_2} {{ t_3}}^{4} +16 {{ t_1}}^{4}-1528 {{ t_1}}^{3}{ t_2}-1528 {{
        t_1}}^{3}{ t_3} -3183 {{ t_1}}^{2}{{ t_2}}^{2}-3182 {{ t_1}}^{2}{ t_2} {
      t_3}-3183 {{ t_1}}^{2}{{ t_3}}^{2}-1528 { t_1} {{ t_2}}^ {3}\\
    -3182 { t_1} {{ t_2}}^{2}{ t_3}-3182 { t_1} { t_2} { { t_3}}^{2} -1528 {
      t_1} {{ t_3}}^{3}+16 {{ t_2}}^{4}-1528 {{ t_2}}^{3}{ t_3}-3183 {{
        t_2}}^{2}{{ t_3}}^{2}-1528 { t_2} {{ t_3}}^{3}+16 {{ t_3}}^{4}\\
    -544 {{t_1}}^{3} -1352 {{ t_1}}^{2}{ t_2}-1352 {{ t_1}}^{2}{ t_3}-1352 {
      t_1} {{ t_2}}^{2}-1656 { t_1} { t_2} { t_3}-1352 { t_1} {{ t_3}}^{2}-544
    {{ t_2}}^{3}-1352 {{ t_2}}^{2}{ t_3}\\
    -1352 { t_2} {{ t_3}}^{2}-544 {{t_3}}^{3} -272 {{ t_1}}^{2}-520 { t_1} {
      t_2}-520 { t_1} { t_3}-272 {{ t_2}}^{2}-520 { t_2 } { t_3}-272 {{
        t_3}}^{2}+32 { t_1}\\
    +32 { t_2}+32 { t_3} +144
  \end{gather*}
\end{small}
which defines a variety in joint space displayed in Fig.~\ref{fig:mokey}.

Now surfaces containing all input singularities of the translational operation
mode of the 3-\underline{R}UU PM have been determined in the space of Study parameters and
in joint space as well as all surfaces containing all output singularities in
the space of Study parameters. In joint space, the simpler of two varieties
containing all output singularities has been determined. By intersecting these
varieties one can easily find poses where the manipulator is simultaneously
input and output singular. An admittedly impractical pose being both input and
output singular is shown in Fig.~\ref{fig:I+Osing}.

\subsection{Constraint Singularities}

Constraint singularities are defined in \cite{bonev2001}. Here they occur when
the Jacobian $\mathbf{J}$ is rank defficient. This is the case if the
determinants of all $8\times 8$ sub-matrices of $\mathbf{J}$ vanish. Since
$\mathbf{J_o}$ is one of these sub-matrices, constraint singularities are in
particular output singularities for this type of manipulator. Geometrically, the
tangent space of the constraint variety gains dimensions in constraint
singularities. Neglecting multiple coverings, these are the points, where
different irreducible components intersect together with their
self-intersections. Constraint singularities are not discussed in more detail
since they are covered by the output singularity surface here, which need to be
avoided in several kinematic tasks anyway, e.g. path planning.

\begin{figure}[h!]
  \centering
  {\includegraphics[width=0.7\linewidth]{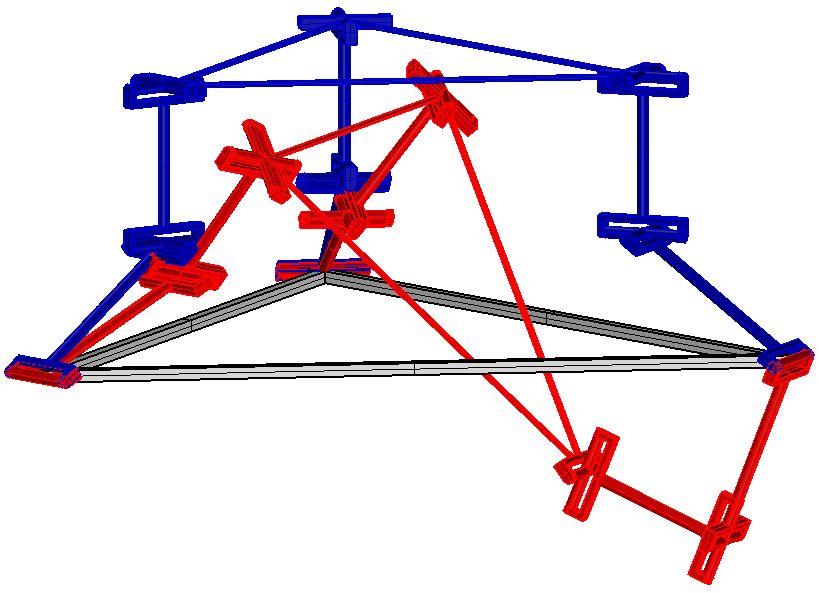}}
  \caption{PM moving out of translational operation mode}\label{fig:change_of_OM}
\end{figure}

Nonetheless it would be desirable to know if the 3-\underline{R}UU PM can switch
from the translational operation mode into another mode. To answer this
question, one can check whether the components $O_1$ and $O_3$ have a non-empty
intersection and thus a change between the respective operation modes is
possible. Since there is no defining ideal for $O_3$ available it is difficult
to compute this intersection. However, by a detailed inspection of the PM, a
real curve $C(t) = [x_0(t), x_1(t), x_2(t), x_3(t), y_0(t), y_1(t), y_2(t),
y_3(t)]$ can be constructed to be fully contained in the workspace. The geometry
of the manipulator suggests, that rotations around the $y$-axis ($x_0(t)=1$,
$x_1(t)=0$, $x_3(t)=0$) as well as translations parallel to the $xz$-plane
($y_0(t)=0$, $y_1(t)=0$) are possible. To further simplify the constraints,
$y_2(t)$ is chosen to be zero. Setting $x_2(t) = t$, one can find $y_3(t)$ such
that the curve lies on the workspace. For the design parameters
\begin{align}\label{eq:pars2}
  a_1= 5,~a_3= 4,~r_0=11,~r_1=7,
\end{align}
above's procedure yields the real curve
\begin{equation}\label{curvec}
\begin{aligned}
  C : [-1,1] &\longrightarrow \PS \\
  t &\longmapsto [1,0,t,0,0,0,0,h / 4],
\end{aligned}
\end{equation}
where
\begin{equation*}
  h \coloneqq 18 t+\sqrt {25 {t}^{2}+36}+\sqrt {{\frac { \left( -59 {t}^{4}+212 
{t}^{2}+256 \right) \sqrt {25 {t}^{2}+36}-600 {t}^{5}-864 {t}^{3}}{
\sqrt {25 {t}^{2}+36} \left( {t}^{2}+4 \right) }}}.
\end{equation*}
It intersects the translational three-space $O_1$ only once, namely at the
parameter value $t = 0$. Even though this curve and its derivation is not of
particular interest it does however provide a proof by example showing that the
intersection $O_1 \cap O_3$ is not empty. Thus the manipulator can indeed change
between its operation modes. Note that because of symmetry, the same holds true
for $\pm 120^{\circ}$ rotations about the $z$-axis of the motion induced by $C$.
Fig.~\ref{fig:change_of_OM} shows two poses of the motion parameterized by $C$.

\subsection{Self-motions}\label{sec:selfmotion}

\begin{figure}[h!]
  \centering
  {\includegraphics[width=0.72\linewidth]{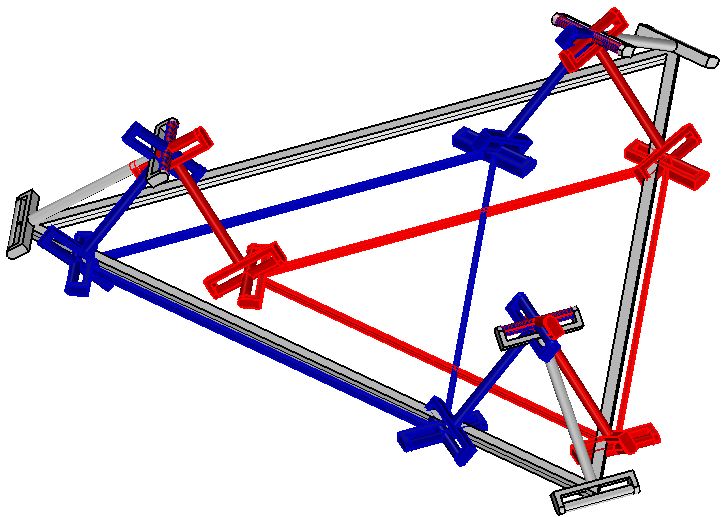}}
  \caption{Two poses of the self-motion}\label{fig:selfmotion}
\end{figure}

A motion the manipulator can perform even though the actuated joints are locked,
is called self-motion. Such a motion can be represented by a curve on the
constraint variety with fixed input parameters. Tangents of this curve are
independent of input parameters and are obviously contained in the tangent space
of the constraint variety. Thus, the tangent space contains elements independent
of the input parameters. Consequently, all points of such a curve are output
singularities. Indeed, a self-motion was found in Section~\ref{sec:output} for
the specific design in Eq.~\eqref{eq:pars}. In the following, conditions on the
design parameters of a 3-\underline{R}UU PM will be derived, for which the
manipulator has such a self-motion. Motivated by the results above, one is
interested to calculate the intersection of $\mathbb{V}(F_1)$ and
$\mathbb{V}(F_2)$. These varieties intersect in a circle contained in the plane
$y_3=0$. The fact that the intersection is contained in this plane suggests to
set the Study parameter $y_3$ to zero. Differentiating the general constraint
equations for the translational operation mode $\mathcal{W_T}$ yields a $3\times
3$ Jacobian depending on $y_1$, $y_2$, $t_1$, $t_2$ and $t_3$. Similar to the
procedure used in Section~\ref{sec:output}, the determinant of this matrix
together with $\mathcal{W_T}$ yields a system of four equations. By successively
eliminating the input parameters via computing resultants one obtains a
polynomial expression, which has the factor
\begin{align*}
  {{ a_1}}^{2}-{{ a_3}}^{2}-(r_0-r_1)^2+4 \left({{ y_1}}^{2}+ {{ y_2}}^{2}\right).
\end{align*}
If one chooses design parameters such that ${a_1}^{2}-{a_3}^{2}-(r_0-r_1)^2$ is
negative, this factor defines a real circle. An easy computation shows, that
every point on this circle can be reached with the fixed input parameters
\begin{align*}
  t_i= {\frac
  {-{ a_1}\pm\sqrt {{{ a_1}}^{2}-{{ r_0}}^{2}+2{ r_0}{ r_1}-{{ r_1}}^{2}}}{{
  r_0}-{ r_1}}},
\end{align*}
provided $r_0\neq r_1$ or $t_i=0$ in case $r_0=r_1$ for $i=1$, $2$, $3$ (compare to
Eq.~\eqref{eq:circle-cond}). Fig.~\ref{fig:selfmotion} shows two poses of this
self-motion for the manipulator given by the design parameters in
Eq.~\eqref{eq:pars2}.

\section{Conclusion}

A detailed singularity analysis of the translational operation mode of the
3-\underline{R}UU parallel manipulator was given. Using a system of algebraic
constraint equations, varieties containing input and output singularities could
be determined in the Study space and in the joint space. Furthermore an input
parameter free characterization of the overall workspace of the manipulator was
found, to the best of the authors knowledge, for the first time. Numerical
computations suggested that the workspace decomposes into three varieties of
dimension three. Two of them are simple but unfortunately one of them seems to
have impractically high degree. However, a curve on this component was found
that intersects one of the simple components in a single point, thus transition
between operation modes is possible. Finally, for this type of PM a self-motion
was discovered. Future research should focus on the singularities of the general
operation mode and on finding all poses of the end effector where the
manipulator can switch between the operation modes.

\section*{Acknowledgements}
This work was conducted with the support of the University of Innsbruck and the
support of the FWF projects KAPAMAT (I 1750-N26), P~31061 and EKIMAP
(P~30673-N32).

\section*{Appendix}

\begin{small}
  \begin{align*}
    s_1 : &  18{{ r_0}}^{2}{{ t_1}}^{2}{{ t_2}}^{2}{ y_3}+18{{ r_0}}^
            {2}{{ t_1}}^{2}{{ t_3}}^{2}{ y_3}+18{{ r_0}}^{2}{{ t_2}}^{
            2}{{ t_3}}^{2}{ y_3}+18{{ r_1}}^{2}{{ t_1}}^{2}{{ t_2}}^{2
            }{ y_3}+18{{ r_1}}^{2}{{ t_1}}^{2}{{ t_3}}^{2}{ y_3}\\
          &+18{{ r_1}}^{2}{{ t_2}}^{2}{{ t_3}}^{2}{ y_3}+24{{ a_1}}^{2}{
            t_1}{ t_2}{ y_3}+24{{ a_1}}^{2}{ t_1}{ t_3}{
            y_3}+24{{ a_1}}^{2}{ t_2}{ t_3}{ y_3}-36{ r_0}{
            r_1}{{ t_1}}^{2}{ y_3}-36{ r_0}{ r_1}{{ t_2}}^{2}
            { y_3}\\
          &-36{ r_0}{ r_1}{{ t_3}}^{2}{ y_3}+24{ a_1}
            { r_0}{ t_1}{ y_3}+24{ a_1}{ r_0}{ t_2}{ y_3
            }+24{ a_1}{ r_0}{ t_3}{ y_3}-24{ a_1}{ r_1}{
            t_1}{ y_3}-24{ a_1}{ r_1}{ t_2}{ y_3}\\
          &-24{a_1}{ r_1}{ t_3}{ y_3}+12{{ a_1}}^{3}{ t_1}{ t_3}
            +12{{ a_1}}^{3}{ t_2}{ t_3}-6{ a_1}{ r_0}{ r_1}
            {{ t_3}}^{2}-3{ a_1}{{ r_0}}^{2}{{ t_1}}^{2}{{ t_3}}^{
            2}-3{ a_1}{{ r_0}}^{2}{{ t_2}}^{2}{{ t_3}}^{2}\\
          &-3{ a_1}{{ r_1}}^{2}{{ t_1}}^{2}{{ t_3}}^{2}-3{ a_1}{{ r_1}}^{
            2}{{ t_2}}^{2}{{ t_3}}^{2}+12{ a_1}{ r_0}{{ t_1}}^{2}{
            { t_2}}^{2}{ y_2}\sqrt {3}-12{ a_1}{ r_0}{{ t_1}}^{2
            }{{ t_3}}^{2}{ y_2}\sqrt {3}\\
          &-12{ a_1}{ r_1}{{ t_1}}^{2}{{ t_2}}^{2}{ y_2}\sqrt {3}+12{ a_1}{ r_1}{{ t_1}
            }^{2}{{ t_3}}^{2}{ y_2}\sqrt {3}-36{ r_0}{ r_1}{{ 
            t_1}}^{2}{{ t_2}}^{2}{{ t_3}}^{2}{ y_3}+24{ a_1}{ r_0}{
            { t_1}}^{2}{{ t_2}}^{2}{ t_3}{ y_3}\\
          &+24{ a_1}{ r_0}{{ t_1}}^{2}{ t_2}{{ t_3}}^{2}{ y_3}+24{ a_1}{ r_0}
            { t_1}{{ t_2}}^{2}{{ t_3}}^{2}{ y_3}-24{ a_1}{ r_1
            }{{ t_1}}^{2}{{ t_2}}^{2}{ t_3}{ y_3}-24{ a_1}{ 
            r_1}{{ t_1}}^{2}{ t_2}{{ t_3}}^{2}{ y_3}\\
          &-24{ a_1}{r_1}{ t_1}{{ t_2}}^{2}{{ t_3}}^{2}{ y_3}+18{{ r_0}}
            ^{2}{ y_3}+18{{ r_1}}^{2}{ y_3}+12{ a_1}{ r_0}{{ 
            t_2}}^{2}{ y_2}\sqrt {3}-12{ a_1}{ r_0}{{ t_3}}^{2}{
            y_2}\sqrt {3}\\
          &-12{ a_1}{ r_1}{{ t_2}}^{2}{ y_2}\sqrt {3}+12{ a_1}{ r_1}{{ t_3}}^{2}{ y_2}\sqrt {3}+8
            {{ a_1}}^{2}{{ t_1}}^{2}{ t_2}{ y_2}\sqrt {3}-8{{ 
            a_1}}^{2}{{ t_1}}^{2}{ t_3}{ y_2}\sqrt {3}+16{{ a_1}}^{2}
            \sqrt {3}{ t_1}{{ t_2}}^{2}{ y_2}\\
          &-16{{ a_1}}^{2}\sqrt {3}{t_1}{{ t_3}}^{2}{ y_2}+8{{ a_1}}^{2}{ t_3}{{ t_2}}^
            {2}{ y_2}\sqrt {3}+18{ a_1}{ r_0}{ r_1}{{ t_1}}^{2
            }{{ t_2}}^{2}{{ t_3}}^{2}+9{ a_1}{{ r_1}}^{2}+12{{ a_1
            }}^{3}{ t_1}{ t_2}\\
          &+6{ a_1}{ r_0}{ r_1}{{ t_1}}^{2}{{ t_2}}^{2}+6{ a_1}{ r_0}{ r_1}{{ t_1}}^{2}{{ 
            t_3}}^{2}-12{{ a_1}}^{2}{ r_0}{{ t_1}}^{2}{{ t_2}}^{2}{ 
            t_3}-12{{ a_1}}^{2}{ r_0}{{ t_1}}^{2}{ t_2}{{ t_3}}^{2}
            -12{{ a_1}}^{2}{ r_0}{ t_1}{{ t_2}}^{2}{{ t_3}}^{2}\\
          &+6{ a_1}{ r_0}{ r_1}{{ t_2}}^{2}{{ t_3}}^{2}+9{ 
            a_1}{{ r_0}}^{2}-12{{ a_1}}^{2}{ r_1}{ t_2}-12{{ a_1}
            }^{2}{ r_1}{ t_1}+12{{ a_1}}^{2}{ r_1}{{ t_1}}^{2}{{
            t_2}}^{2}{ t_3}+12{{ a_1}}^{2}{ r_1}{{ t_1}}^{2}{ t_2
            }{{ t_3}}^{2}\\
          &+12{{ a_1}}^{2}{ r_1}{ t_1}{{ t_2}}^{2}{{ t_3}}^{2}+18{{ r_0}}^{2}{{ t_1}}^{2}{{ t_2}}^{2}
            {{ t_3}}^{2}{ y_3}+18{{ r_1}}^{2}{{ t_1}}^{2}{{ t_2}}^{2}{{ t_3}}
            ^{2}{ y_3}+24{{ a_1}}^{2}{{ t_1}}^{2}{ t_2}{ t_3}{ 
            y_3}+24{{ a_1}}^{2}{ t_1}{{ t_2}}^{2}{ t_3}{ y_3}\\
          &+24{{ a_1}}^{2}{ t_1}{ t_2}{{ t_3}}^{2}{ y_3}-36{ r_0}
            { r_1}{{ t_1}}^{2}{{ t_2}}^{2}{ y_3}-36{ r_0}{ r_1}
            {{ t_1}}^{2}{{ t_3}}^{2}{ y_3}-36{ r_0}{ r_1}{{ 
            t_2}}^{2}{{ t_3}}^{2}{ y_3}+24{ a_1}{ r_0}{{ t_1}}^{2}{
            t_2}{ y_3}\\
          &+24{ a_1}{ r_0}{{ t_1}}^{2}{ t_3}{y_3}+24{ a_1}{ r_0}{ t_1}{{ t_2}}^{2}{ y_3}+24{
            a_1}{ r_0}{ t_1}{{ t_3}}^{2}{ y_3}+24{ a_1}{
            r_0}{{ t_2}}^{2}{ t_3}{ y_3}+24{ a_1}{ r_0}{
            t_2}{{ t_3}}^{2}{ y_3}\\
          &-24{ a_1}{ r_1}{{ t_1}}^{2}{ t_2}{ y_3}-24{ a_1}{ r_1}{{ t_1}}^{2}{ t_3}{
            y_3}-24{ a_1}{ r_1}{ t_1}{{ t_2}}^{2}{ y_3}-24{
            a_1}{ r_1}{ t_1}{{ t_3}}^{2}{ y_3}-24{ a_1}{
            r_1}{{ t_2}}^{2}{ t_3}{ y_3}\\
          &-24{ a_1}{ r_1}{t_2}{{ t_3}}^{2}{ y_3}+18{{ r_0}}^{2}{{ t_1}}^{2}{ 
            y_3}+18{{ r_0}}^{2}{{ t_2}}^{2}{ y_3}+18{{ r_0}}^{2}{{ 
            t_3}}^{2}{ y_3}+18{{ r_1}}^{2}{{ t_1}}^{2}{ y_3}+18{{ r_1
            }}^{2}{{ t_2}}^{2}{ y_3}\\
          &+18{{ r_1}}^{2}{{ t_3}}^{2}{ y_3}-36{ r_0}{ r_1}{ y_3}-12{ a_1}{ r_1}{{ t_1}}^{2
            }{{ t_2}}^{2}{ y_1}-12{ a_1}{ r_1}{{ t_1}}^{2}{{ t_3
            }}^{2}{ y_1}+24{ a_1}{ r_1}{{ t_2}}^{2}{{ t_3}}^{2}{
            y_1}\\
          &+12{ a_1}{ r_0}{{ t_1}}^{2}{{ t_2}}^{2}{ y_1}+
            12{ a_1}{ r_0}{{ t_1}}^{2}{{ t_3}}^{2}{ y_1}-24{ 
            a_1}{ r_0}{{ t_2}}^{2}{{ t_3}}^{2}{ y_1}+3{ a_1}{{
            r_0}}^{2}{{ t_1}}^{2}+3{ a_1}{{ r_0}}^{2}{{ t_2}}^{2}+3
            { a_1}{{ r_0}}^{2}{{ t_3}}^{2}\\
          &+3{ a_1}{{ r_1}}^{2}{{t_1}}^{2}+3{ a_1}{{ r_1}}^{2}{{ t_2}}^{2}+3{ a_1}{{
            r_1}}^{2}{{ t_3}}^{2}-9{ a_1}{{ r_0}}^{2}{{ t_1}}^{2}{{
            t_2}}^{2}{{ t_3}}^{2}-9{ a_1}{{ r_1}}^{2}{{ t_1}}^{2}{{
            t_2}}^{2}{{ t_3}}^{2}-18{ a_1}{ r_0}{ r_1}\\
          &+12{{a_1}}^{2}{ r_0}{ t_1}+12{{ a_1}}^{2}{ r_0}{ t_2}+12{{
            a_1}}^{2}{ r_0}{ t_3}-12{{ a_1}}^{2}{ r_1}{ t_3}+24
            {{ a_1}}^{2}{{ t_1}}^{2}{ t_2}{ y_1}+24{{ a_1}}^{2}{{
            t_1}}^{2}{ t_3}{ y_1}\\
          &-24{{ a_1}}^{2}{{ t_2}}^{2}{ t_3}{ y_1}-24{{ a_1}}^{2}{ t_2}{{ t_3}}^{2}{ y_1}+24{
            a_1}{ r_0}{{ t_1}}^{2}{ y_1}-12{ a_1}{ r_0}{{
            t_2}}^{2}{ y_1}-12{ a_1}{ r_0}{{ t_3}}^{2}{ y_1}-24
            { a_1}{ r_1}{{ t_1}}^{2}{ y_1}\\
          &+12{ a_1}{ r_1}{{ t_2}}^{2}{ y_1}+12{ a_1}{ r_1}{{ t_3}}^{2}{ y_1}-8
            {{ a_1}}^{2}{ t_2}{{ t_3}}^{2}{ y_2}\sqrt {3}-3{ a_1
            }{{ r_0}}^{2}{{ t_1}}^{2}{{ t_2}}^{2}-3{ a_1}{{ r_1}}^
            {2}{{ t_1}}^{2}{{ t_2}}^{2}-6{ a_1}{ r_0}{ r_1}{{
            t_1}}^{2}\\
          &-6{ a_1}{ r_0}{ r_1}{{ t_2}}^{2}-12{{
            a_1}}^{3}{{ t_1}}^{2}{ t_2}{ t_3}-12{{ a_1}}^{3}{ t_1
            }{{ t_2}}^{2}{ t_3}-12{{ a_1}}^{3}{ t_1}{ t_2}{{ 
            t_3}}^{2}
  \end{align*}
\end{small}

\begin{small}
  \begin{align*}
    s_2 : &{ a_1}{{ t_1}}^{2}{{ t_2}}^{2}{{ y_1}}^{2}+{ a_1}{{ 
            t_1}}^{2}{{ t_3}}^{2}{{ y_1}}^{2}-5{ a_1}{{ t_2}}^{2}{{ 
            t_3}}^{2}{{ y_1}}^{2}-6{{ r_0}}^{2}{{ t_1}}^{2}{{ t_2}}^{2}{
            y_3}-6{{ r_0}}^{2}{{ t_1}}^{2}{{ t_3}}^{2}{ y_3}-6{{
            r_0}}^{2}{{ t_2}}^{2}{{ t_3}}^{2}{ y_3}\\
          &-6{{ r_1}}^{2}{{t_1}}^{2}{{ t_2}}^{2}{ y_3}-6{{ r_1}}^{2}{{ t_1}}^{2}{{
            t_3}}^{2}{ y_3}-6{{ r_1}}^{2}{{ t_2}}^{2}{{ t_3}}^{2}{
            y_3}-8{{ a_1}}^{2}{ t_1}{ t_2}{ y_3}-8{{ a_1}}^{2
            }{ t_1}{ t_3}{ y_3}-8{{ a_1}}^{2}{ t_2}{ t_3}{
            y_3}\\
          &+12{ r_0}{ r_1}{{ t_1}}^{2}{ y_3}+12{ r_0}{
            r_1}{{ t_2}}^{2}{ y_3}+12{ r_0}{ r_1}{{ t_3}}^{2}
            { y_3}-8{ a_1}{ r_0}{ t_1}{ y_3}-8{ a_1}{ 
            r_0}{ t_2}{ y_3}-8{ a_1}{ r_0}{ t_3}{ y_3}\\
          &+8{a_1}{ r_1}{ t_1}{ y_3}+8{ a_1}{ r_1}{ t_2}
            { y_3}+8{ a_1}{ r_1}{ t_3}{ y_3}+6{{ t_1}}^{2}
            {{ t_2}}^{2}{{ y_1}}^{2}{ y_3}+6{{ t_1}}^{2}{{ t_2}}^{2}{{
            y_2}}^{2}{ y_3}+6{{ t_1}}^{2}{{ t_3}}^{2}{{ y_1}}^{2}{
            y_3}\\
          &+6{{ t_1}}^{2}{{ t_3}}^{2}{{ y_2}}^{2}{ y_3}+6{{
            t_2}}^{2}{{ t_3}}^{2}{{ y_1}}^{2}{ y_3}+6{{ t_2}}^{2}{{
            t_3}}^{2}{{ y_2}}^{2}{ y_3}+8{ a_1}{ t_1}{ y_1}{
            y_3}-4{ a_1}{ t_2}{ y_1}{ y_3}-4{ a_1}{ t_3
            }{ y_1}{ y_3}\\
          &-4{{ a_1}}^{3}{ t_1}{ t_3}-4{{ a_1}
            }^{3}{ t_2}{ t_3}+2{ a_1}{ r_0}{ r_1}{{ t_3}}^{2
            }+{ a_1}{{ r_0}}^{2}{{ t_1}}^{2}{{ t_3}}^{2}+{ a_1}{{
            r_0}}^{2}{{ t_2}}^{2}{{ t_3}}^{2}+{ a_1}{{ r_1}}^{2}{{
            t_1}}^{2}{{ t_3}}^{2}+{ a_1}{{ r_1}}^{2}{{ t_2}}^{2}{{
            t_3}}^{2}\\
          &+2{{ a_1}}^{2}{{ t_1}}^{2}{ t_3}{{ t_2}}^{2}{
            y_2}\sqrt {3}-2{{ a_1}}^{2}{{ t_1}}^{2}{ t_2}{{ t_3}}
            ^{2}{ y_2}\sqrt {3}-2{ a_1}{ r_0}{{ t_1}}^{2}{{ t_2}
            }^{2}{ y_2}\sqrt {3}+2{ a_1}{ r_0}{{ t_1}}^{2}{{ t_3
            }}^{2}{ y_2}\sqrt {3}\\
          &+2{ a_1}{ r_1}{{ t_1}}^{2}{{t_2}}^{2}{ y_2}\sqrt {3}-2{ a_1}{ r_1}{{ t_1}}^{2}{{
            t_3}}^{2}{ y_2}\sqrt {3}-4{ a_1}{{ t_1}}^{2}{{ t_2}}^
            {2}{ y_1}{ y_2}\sqrt {3}+4{ a_1}{{ t_1}}^{2}{{ t_3}}
            ^{2}{ y_1}{ y_2}\sqrt {3}\\
          &+4{ a_1}{{ t_1}}^{2}\sqrt {3}
            { t_2}{ y_2}{ y_3}-4{ a_1}{{ t_1}}^{2}\sqrt {3}{ 
            t_3}{ y_2}{ y_3}-4{ a_1}{{ t_2}}^{2}\sqrt {3}{ t_3}
            { y_2}{ y_3}+4{ a_1}\sqrt {3}{ t_2}{{ t_3}}^{2}{ 
            y_2}{ y_3}+2{{ a_1}}^{2}{ t_2}{ y_2}\sqrt {3}\\
          &-2{{a_1}}^{2}{ t_3}{ y_2}\sqrt {3}+12{ r_0}{ r_1}{{ t_1
            }}^{2}{{ t_2}}^{2}{{ t_3}}^{2}{ y_3}-8{ a_1}{ r_0}{{
            t_1}}^{2}{{ t_2}}^{2}{ t_3}{ y_3}-8{ a_1}{ r_0}{{
            t_1}}^{2}{ t_2}{{ t_3}}^{2}{ y_3}-8{ a_1}{ r_0}{
            t_1}{{ t_2}}^{2}{{ t_3}}^{2}{ y_3}\\
          &+8{ a_1}{ r_1}{{ t_1}}^{2}{{ t_2}}^{2}{ t_3}{ y_3}+8{ a_1}{ r_1}{
            { t_1}}^{2}{ t_2}{{ t_3}}^{2}{ y_3}+8{ a_1}{ r_1}{
            t_1}{{ t_2}}^{2}{{ t_3}}^{2}{ y_3}-4{ a_1}{{ t_1}}^
            {2}{{ t_2}}^{2}{ t_3}{ y_1}{ y_3}-4{ a_1}{{ t_1}}^
            {2}{ t_2}{{ t_3}}^{2}{ y_1}{ y_3}\\
          &+8{ a_1}{ t_1}{{ t_2}}^{2}{{ t_3}}^{2}{ y_1}{ y_3}-4{ a_1}{{ t_1}}^
            {2}{{ t_2}}^{2}\sqrt {3}{ t_3}{ y_2}{ y_3}+4{ a_1}{{
            t_1}}^{2}\sqrt {3}{ t_2}{{ t_3}}^{2}{ y_2}{ y_3}+3{
            a_1}{{ y_1}}^{2}-6{{ r_0}}^{2}{ y_3}-6{{ r_1}}^{2}{
            y_3}\\
          &+6{{ y_1}}^{2}{ y_3}+6{{ y_2}}^{2}{ y_3}-2{ a_1
            }{ r_0}{{ t_2}}^{2}{ y_2}\sqrt {3}+2{ a_1}{ r_0}
            {{ t_3}}^{2}{ y_2}\sqrt {3}+2{ a_1}{ r_1}{{ t_2}}^{2
            }{ y_2}\sqrt {3}-2{ a_1}{ r_1}{{ t_3}}^{2}{ y_2}
            \sqrt {3}\\
          &-4{ a_1}{{ t_2}}^{2}{ y_1}{ y_2}\sqrt {3}+4
            { a_1}{{ t_3}}^{2}{ y_1}{ y_2}\sqrt {3}+4{ a_1}
            \sqrt {3}{ t_2}{ y_2}{ y_3}-4{ a_1}\sqrt {3}{ t_3}
            { y_2}{ y_3}-2{{ a_1}}^{2}{{ t_1}}^{2}{ t_2}{ y_2}
            \sqrt {3}\\
          &+2{{ a_1}}^{2}{{ t_1}}^{2}{ t_3}{ y_2}\sqrt {3}
            -4{{ a_1}}^{2}\sqrt {3}{ t_1}{{ t_2}}^{2}{ y_2}+4{{ 
            a_1}}^{2}\sqrt {3}{ t_1}{{ t_3}}^{2}{ y_2}-2{{ a_1}}^{2}{
            t_3}{{ t_2}}^{2}{ y_2}\sqrt {3}-6{ a_1}{ r_0}{
            r_1}{{ t_1}}^{2}{{ t_2}}^{2}{{ t_3}}^{2}\\
          &-3{ a_1}{{r_1}}^{2}-4{{ a_1}}^{3}{ t_1}{ t_2}-3{ a_1}{{ t_1
            }}^{2}{{ t_2}}^{2}{{ y_2}}^{2}-3{ a_1}{{ t_1}}^{2}{{ t_3
            }}^{2}{{ y_2}}^{2}+3{ a_1}{{ t_2}}^{2}{{ t_3}}^{2}{{ y_2
            }}^{2}+3{ a_1}{{ y_2}}^{2}\\
          &-3{ a_1}{{ t_1}}^{2}{{ t_2
            }}^{2}{{ t_3}}^{2}{{ y_1}}^{2}+2{{ a_1}}^{2}{{ t_1}}^{2}{{
            t_2}}^{2}{ t_3}{ y_1}+2{{ a_1}}^{2}{{ t_1}}^{2}{ t_2}
            {{ t_3}}^{2}{ y_1}-4{{ a_1}}^{2}{ t_1}{{ t_2}}^{2}{{
            t_3}}^{2}{ y_1}-2{ a_1}{ r_0}{ r_1}{{ t_1}}^{2}{{
            t_2}}^{2}\\
          &-2{ a_1}{ r_0}{ r_1}{{ t_1}}^{2}{{ t_3}}
            ^{2}+4{{ a_1}}^{2}{ r_0}{{ t_1}}^{2}{{ t_2}}^{2}{ t_3}+4
            {{ a_1}}^{2}{ r_0}{{ t_1}}^{2}{ t_2}{{ t_3}}^{2}+4{{
            a_1}}^{2}{ r_0}{ t_1}{{ t_2}}^{2}{{ t_3}}^{2}-3{ 
            a_1}{{ t_1}}^{2}{{ t_2}}^{2}{{ t_3}}^{2}{{ y_2}}^{2}\\
          &-2{a_1}{ r_0}{ r_1}{{ t_2}}^{2}{{ t_3}}^{2}-3{ a_1}{{
            r_0}}^{2}+4{{ a_1}}^{2}{ r_1}{ t_2}+4{{ a_1}}^{2}{
            r_1}{ t_1}-4{{ a_1}}^{2}{ r_1}{{ t_1}}^{2}{{ t_2}}^
            {2}{ t_3}-4{{ a_1}}^{2}{ r_1}{{ t_1}}^{2}{ t_2}{{ 
            t_3}}^{2}\\
          &-4{{ a_1}}^{2}{ r_1}{ t_1}{{ t_2}}^{2}{{ t_3}}
            ^{2}-6{{ r_0}}^{2}{{ t_1}}^{2}{{ t_2}}^{2}{{ t_3}}^{2}{ 
            y_3}-6{{ r_1}}^{2}{{ t_1}}^{2}{{ t_2}}^{2}{{ t_3}}^{2}{ y_3
            }-8{{ a_1}}^{2}{{ t_1}}^{2}{ t_2}{ t_3}{ y_3}-8{{
            a_1}}^{2}{ t_1}{{ t_2}}^{2}{ t_3}{ y_3}\\
          &-8{{ a_1}}^{2}{ t_1}{ t_2}{{ t_3}}^{2}{ y_3}+12{ r_0}{ r_1}{
            { t_1}}^{2}{{ t_2}}^{2}{ y_3}+12{ r_0}{ r_1}{{ t_1}}
            ^{2}{{ t_3}}^{2}{ y_3}+12{ r_0}{ r_1}{{ t_2}}^{2}{{
            t_3}}^{2}{ y_3}-8{ a_1}{ r_0}{{ t_1}}^{2}{ t_2}{
            y_3}\\
          &-8{ a_1}{ r_0}{{ t_1}}^{2}{ t_3}{ y_3}-8{
            a_1}{ r_0}{ t_1}{{ t_2}}^{2}{ y_3}-8{ a_1}{ 
            r_0}{ t_1}{{ t_3}}^{2}{ y_3}-8{ a_1}{ r_0}{{ t_2}
            }^{2}{ t_3}{ y_3}-8{ a_1}{ r_0}{ t_2}{{ t_3}}^{2
            }{ y_3}+8{ a_1}{ r_1}{{ t_1}}^{2}{ t_2}{ y_3}\\
          &+8{a_1}{ r_1}{{ t_1}}^{2}{ t_3}{ y_3}+8{ a_1}{ 
            r_1}{ t_1}{{ t_2}}^{2}{ y_3}+8{ a_1}{ r_1}{ t_1}
            {{ t_3}}^{2}{ y_3}+8{ a_1}{ r_1}{{ t_2}}^{2}{ t_3}
            { y_3}+8{ a_1}{ r_1}{ t_2}{{ t_3}}^{2}{ y_3}\\
          &+6{{ t_1}}^{2}{{ t_2}}^{2}{{ t_3}}^{2}{{ y_1}}^{2}{ y_3}+6{{
            t_1}}^{2}{{ t_2}}^{2}{{ t_3}}^{2}{{ y_2}}^{2}{ y_3}-4{
            a_1}{{ t_1}}^{2}{ t_2}{ y_1}{ y_3}-4{ a_1}{{
            t_1}}^{2}{ t_3}{ y_1}{ y_3}+8{ a_1}{ t_1}{{ 
            t_2}}^{2}{ y_1}{ y_3}\\
          &+8{ a_1}{ t_1}{{ t_3}}^{2}{ 
            y_1}{ y_3}-4{ a_1}{{ t_2}}^{2}{ t_3}{ y_1}{ y_3}-
            4{ a_1}{ t_2}{{ t_3}}^{2}{ y_1}{ y_3}+6{{ t_1}}^
            {2}{{ y_1}}^{2}{ y_3}+6{{ t_1}}^{2}{{ y_2}}^{2}{ y_3}+6{
            { t_2}}^{2}{{ y_1}}^{2}{ y_3}\\
          &+6{{ t_2}}^{2}{{ y_2}}^{2}{
            y_3}+6{{ t_3}}^{2}{{ y_1}}^{2}{ y_3}+6{{ t_3}}^{2}{{
            y_2}}^{2}{ y_3}-6{{ r_0}}^{2}{{ t_1}}^{2}{ y_3}-6{{ 
            r_0}}^{2}{{ t_2}}^{2}{ y_3}-6{{ r_0}}^{2}{{ t_3}}^{2}{ y_3}
            -6{{ r_1}}^{2}{{ t_1}}^{2}{ y_3}\\
          &-6{{ r_1}}^{2}{{ t_2}}^{
            2}{ y_3}-6{{ r_1}}^{2}{{ t_3}}^{2}{ y_3}+12{ r_0}{ 
            r_1}{ y_3}+2{ a_1}{ r_1}{{ t_1}}^{2}{{ t_2}}^{2}{ 
            y_1}+2{ a_1}{ r_1}{{ t_1}}^{2}{{ t_3}}^{2}{ y_1}-4{
            a_1}{ r_1}{{ t_2}}^{2}{{ t_3}}^{2}{ y_1}\\
          &-2{ a_1}{r_0}{{ t_1}}^{2}{{ t_2}}^{2}{ y_1}-2{ a_1}{ r_0}{
            { t_1}}^{2}{{ t_3}}^{2}{ y_1}+4{ a_1}{ r_0}{{ t_2}}^
            {2}{{ t_3}}^{2}{ y_1}+4{{ a_1}}^{2}{ t_1}{ y_1}-2{{
            a_1}}^{2}{ t_2}{ y_1}-2{{ a_1}}^{2}{ t_3}{ y_1}\\
          &+5{ a_1}{{ t_1}}^{2}{{ y_1}}^{2}-{ a_1}{{ t_2}}^{2}{{ 
            y_1}}^{2}-{ a_1}{{ t_3}}^{2}{{ y_1}}^{2}-{ a_1}{{ r_0}}^{
            2}{{ t_1}}^{2}-{ a_1}{{ r_0}}^{2}{{ t_2}}^{2}-{ a_1}{{
            r_0}}^{2}{{ t_3}}^{2}-{ a_1}{{ r_1}}^{2}{{ t_1}}^{2}-{
            a_1}{{ r_1}}^{2}{{ t_2}}^{2}\\
          &-{ a_1}{{ r_1}}^{2}{{ t_3
            }}^{2}+3{ a_1}{{ r_0}}^{2}{{ t_1}}^{2}{{ t_2}}^{2}{{ t_3
            }}^{2}+3{ a_1}{{ r_1}}^{2}{{ t_1}}^{2}{{ t_2}}^{2}{{ t_3
            }}^{2}+6{ a_1}{ r_0}{ r_1}-4{{ a_1}}^{2}{ r_0}{
            t_1}-4{{ a_1}}^{2}{ r_0}{ t_2}-4{{ a_1}}^{2}{ r_0}
            { t_3}\\
          &+4{{ a_1}}^{2}{ r_1}{ t_3}-6{{ a_1}}^{2}{{ 
            t_1}}^{2}{ t_2}{ y_1}-6{{ a_1}}^{2}{{ t_1}}^{2}{ t_3}{
            y_1}+6{{ a_1}}^{2}{{ t_2}}^{2}{ t_3}{ y_1}+6{{ a_1}
            }^{2}{ t_2}{{ t_3}}^{2}{ y_1}-4{ a_1}{ r_0}{{ t_1}
            }^{2}{ y_1}\\
          &+2{ a_1}{ r_0}{{ t_2}}^{2}{ y_1}+2{ a_1
            }{ r_0}{{ t_3}}^{2}{ y_1}+4{ a_1}{ r_1}{{ t_1}}^
            {2}{ y_1}-2{ a_1}{ r_1}{{ t_2}}^{2}{ y_1}-2{ a_1}
            { r_1}{{ t_3}}^{2}{ y_1}+2{{ a_1}}^{2}{ t_2}{{ t_3
            }}^{2}{ y_2}\sqrt {3}\\
          &+{ a_1}{{ r_0}}^{2}{{ t_1}}^{2}{{ 
            t_2}}^{2}+{ a_1}{{ r_1}}^{2}{{ t_1}}^{2}{{ t_2}}^{2}+2{ 
            a_1}{ r_0}{ r_1}{{ t_1}}^{2}+2{ a_1}{ r_0}{ r_1
            }{{ t_2}}^{2}+4{{ a_1}}^{3}{{ t_1}}^{2}{ t_2}{ t_3}+4
            {{ a_1}}^{3}{ t_1}{{ t_2}}^{2}{ t_3}\\
          &+4{{ a_1}}^{3}{
            t_1}{ t_2}{{ t_3}}^{2}-3{ a_1}{{ t_1}}^{2}{{ y_2}
            }^{2}+3{ a_1}{{ t_2}}^{2}{{ y_2}}^{2}+3{ a_1}{{ t_3}
            }^{2}{{ y_2}}^{2}.
  \end{align*}
\end{small}

\bibliographystyle{alpha}
%\bibliography{biblio}
%\printbibliography

\end{document}